%% file: main.tex
\documentclass{article}
\usepackage{iclr2027_conference,times}

\usepackage[utf8]{inputenc}
\usepackage[T1]{fontenc}
\usepackage{amsmath,amssymb}
\usepackage{graphicx}
\usepackage{booktabs,multirow,array}
\usepackage{nicefrac}
\usepackage{microtype}
\usepackage{xcolor}
\usepackage{float}
\usepackage{enumitem}
\usepackage{hyperref}
\usepackage{url}
\hypersetup{hidelinks}
\newenvironment{contriblist}{\begin{itemize}[leftmargin=*,nosep,itemsep=2pt,topsep=0pt]}{\end{itemize}}

\title{X-MoD: Practical Scaling Laws for Sparse-Depth Routing Beyond Mixture-of-Depths}
\input{authors}
\iclrfinalcopy

\begin{document}
\maketitle
\fancyhead{}
\renewcommand{\headrulewidth}{0pt}
\begin{abstract}
\input{chapters/Abstract}
\end{abstract}

\input{chapters/Introduction}
\input{chapters/Architecture}
\input{chapters/ScalingLaw}
\input{chapters/Experiments}
\input{chapters/Limitations}
\input{chapters/RelatedWorks}
\input{chapters/Conclusion}
\label{page:main_text_end}

\input{chapters/Acknowledgments}
\input{chapters/AIUse}
\bibliographystyle{iclr2027_conference}
\bibliography{references}

\clearpage
\appendix
\input{chapters/Appendix}
\end{document}

%% file: authors.tex
\author{%
\makebox[\dimexpr\textwidth-2\tabcolsep\relax][c]{%
\begin{tabular}[t]{@{}c@{}}
\textbf{Bowen Dong\textsuperscript{1,*}, Yilong Fan\textsuperscript{2,*}, Tengyu Pan\textsuperscript{1}, Yike Zhang\textsuperscript{1}, Zhenyu Li\textsuperscript{1}}\\
\textbf{Zijian Zhang\textsuperscript{2}, Xuewei Li\textsuperscript{2}, Mei Yu\textsuperscript{2}, Jianyong Wang\textsuperscript{1,$\dagger$}}\\[0.5em]
\normalfont\textsuperscript{1}Tsinghua University\quad\textsuperscript{2}Tianjin University\\
\normalfont\small\textsuperscript{*}These authors contributed equally to this work.\\
\normalfont\small\textsuperscript{$\dagger$}Corresponding author: \texttt{jianyong@tsinghua.edu.cn}
\end{tabular}%
}%
}

%% file: chapters/Abstract.tex
Mixture-of-Depths (MoD) enables conditional computation across Transformer depth by routing only a subset of tokens through selected layers, but its original one-sparse--one-dense alternation tightly couples total capacity to active capacity and limits sparse-depth scaling.
We introduce X-MoD, a scalable sparse-depth architecture that decouples token sparsity from anchor stride, allowing total parameter count to grow while keeping active-equivalent capacity nearly fixed.
To make deep sparse routing trainable, X-MoD combines dense anchors with variance-scaled layer-wise gating and depth-wise token balancing.
To make this regime analyzable and usable, we formulate sparse-depth routing as a conditional architecture-design problem: given compute, context length, and active-equivalent backbone size, how should the routing configuration be chosen?
We develop a practical scaling-law framework by fitting X-MoD relative to FLOP-matched dense baselines, yielding an interpretable law that decomposes performance into sparse-capacity gain, sparse-context correction, and anchor-stride interaction.
The law predicts validation loss across routing configurations and reveals how context length, model scale, and anchor stride shape sparse-depth performance.
We validate the architecture and law through pretraining sweeps, held-out scaling-law prediction, ablations, downstream evaluations, and comparisons with Dense, MoD, and representative MoE baselines.\footnote{Anonymous code release: \url{https://anonymous.4open.science/r/X-MoD/}.}

%% file: chapters/Introduction.tex
\section{Introduction}
\label{sec:introduction}

Scaling model capacity has repeatedly proven to be one of the most reliable ways to improve language model performance and unlock new capabilities. 
Classical scaling-law studies for dense Transformers show that model size, data, and training compute interact in a highly structured way, leading to predictable compute-optimal trade-offs \citep{kaplan2020,hoffmann2022}. 
More recently, sparse architectures such as Mixture-of-Experts (MoE) have demonstrated that total parameter count and per-example computation need not grow in lockstep: 
by activating only a subset of parameters for each input, one can substantially increase model capacity without a proportional increase in FLOPs~\citep{MoE2017,GShard2020,SwitchTransformer2022,DeepSeekMoE2024}. 
This raises two natural questions for token routing across depth: 
\emph{Can sparse-depth architectures achieve a similar decoupling between total and active capacities? Can we derive practical scaling laws that guide their design?}

Mixture-of-Depths (MoD) routes selected tokens through Transformer layers; the best-performing configuration in the original study alternates sparse and dense layers \citep{MoD}.
This fundamentally limits sparse-depth scaling: total and active-equivalent parameter counts remain tightly coupled, with a ratio below two for equal-sized layers, preventing the aggressive total-capacity expansion enabled by MoE.
Breaking this limit requires longer sparse stacks, which can concentrate updates on a small token subset; greater token sparsity also reduces attention context.
Scaling sparse depth therefore requires both a trainable architecture and a rule for balancing capacity against context.

In this paper, we introduce \textbf{X-MoD}, a scalable sparse-depth architecture that generalizes MoD beyond strict one-sparse--one-dense alternation. 
X-MoD disentangles two degrees of freedom that are tightly coupled in the original design: 
the token sparsity ratio \(K\), which determines the fraction of tokens entering each sparse layer, and the \emph{anchor stride} \(A\), which controls how many sparse refinements a token receives on average between adjacent dense anchors. 
This decoupling allows X-MoD to substantially increase total parameter count while keeping the active-equivalent capacity nearly fixed. 
To make this regime trainable, we combine dense anchors with three simple but effective mechanisms: 
layer-wise gating with variance scaling, depth-wise token balancing, and an asymmetric hierarchy.

To guide configuration within this expanded design space, we develop a practical scaling-law framework for X-MoD. 
We formulate sparse-depth routing as a conditional architecture-design problem: given a training compute budget, a context length, and an active-equivalent backbone size, how should the architecture be configured---specifically, how should \(K\) and \(A\) be chosen? 
Rather than fitting raw loss directly, we construct a FLOP-matched residual over a dense baseline, which isolates sparse-specific gains and costs while preserving the dense backbone as a stable reference. 
This residual formulation simplifies to a compact and interpretable law that captures how total versus active capacity, sparse-layer context, and anchor stride jointly shape performance, thereby turning sparse-depth routing from a purely architectural heuristic into a quantitatively analyzable design dimension and providing quantitative guidance for configuration selection.

Our main contributions are as follows:
\begin{contriblist}
    \item \textbf{Architecture:} We propose \textbf{X-MoD}, a scalable sparse-depth Transformer that generalizes Mixture-of-Depths through independently controllable token sparsity \(K\) and anchor stride \(A\), enabling a substantially larger gap between total and active-equivalent parameter counts.
    \item \textbf{Scaling law:} We formulate sparse-depth routing as a conditional architecture-design problem and develop a \textbf{practical scaling-law framework} via a FLOP-matched residual over dense baselines, capturing total capacity, sparse-layer context, and anchor stride.
    \item \textbf{Empirical validation:} We validate both the architecture and the resulting law empirically through held-out scaling-law prediction, pilot-selected X-MoD configurations, ablations, downstream evaluations, and comparisons with Dense, MoD, and representative MoE baselines.
\end{contriblist}

%% file: chapters/Architecture.tex
\section{X-MoD: A Scalable Sparse-Depth Architecture}
\label{sec:xmod_architecture}

\begin{figure}[t]
  \centering
  \includegraphics[width=0.85\textwidth]{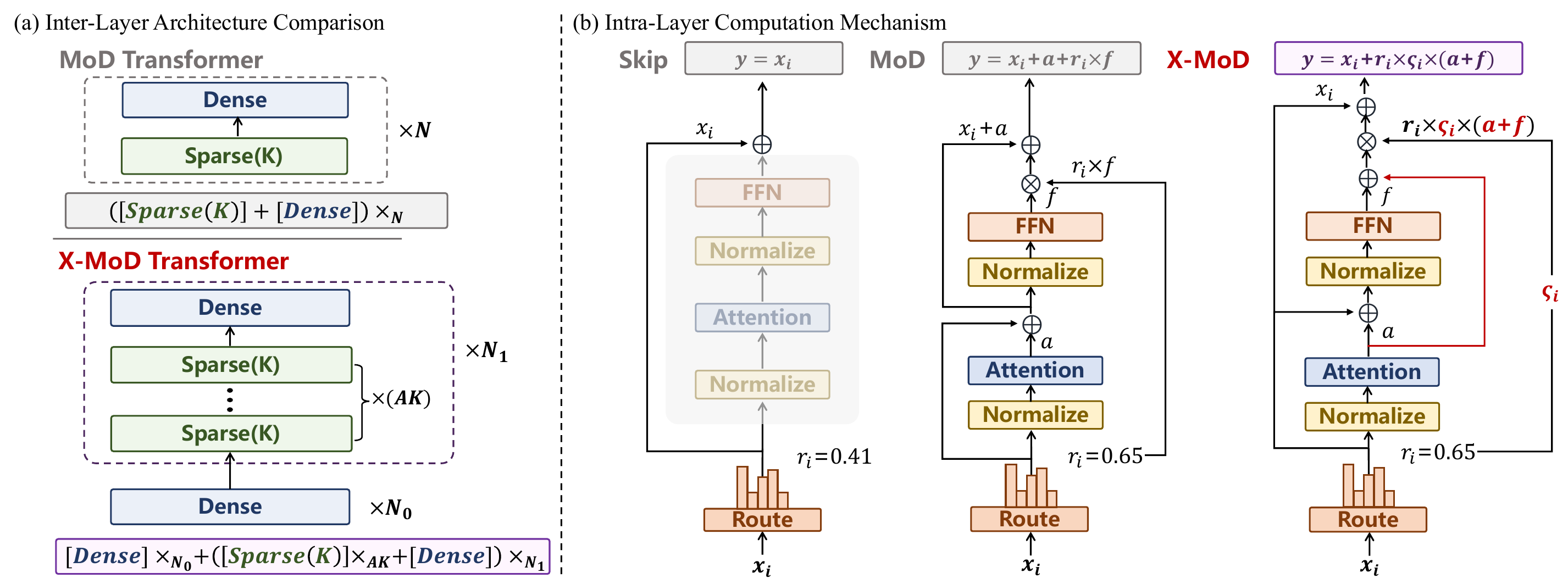}
  \caption{\textbf{MoD vs. X-MoD.} X-MoD decouples token sparsity \(K\) from anchor stride \(A\), enabling a substantially larger gap between total and active-equivalent parameter counts.}
  \label{fig:XMoD}
  \vspace{-12pt}
\end{figure}

\subsection{Revisiting Mixture-of-Depths}
\label{sec:revisiting_mod}

Mixture-of-Depths (MoD) introduces conditional computation along the depth dimension by routing only a subset of tokens through selected Transformer layers \citep{MoD}.
In its original form, the architecture follows a strict dense--sparse alternation:
\begin{equation}
([Sparse(K)] + [Dense])_{\times N},
\label{eq:mod_arch}
\end{equation}
where \( [Sparse(K)] \) denotes a sparse layer that processes only the top-$\frac{1}{K}$ fraction of tokens, while the remaining tokens bypass the layer through an identity path.

This design is effective for stabilizing training: every sparse layer is immediately followed by a dense layer that restores full-token communication and mitigates token imbalance and gradient instability when sparse routing is stacked across depth.
However, the same alternation also imposes a structural limit on how sparse the architecture can become.
Within each sparse--dense pair, the average activated capacity per token is \(1+\frac{1}{K}\), whereas the total parameter capacity is $2$ layers.
Equivalently, the total-to-active parameter ratio is
\begin{equation}
\frac{2}{1+1/K}=\frac{2K}{K+1}<2.
\label{eq:mod_ratio_bound}
\end{equation}
Hence, even when \(K\) is large, the original MoD design cannot substantially decouple total parameter count from active-equivalent parameter count.
In practice, \citet{MoD} report an optimum around \(K=8\) under this restricted regime.

Thus, unlike MoE, the original MoD, with the fixed one-sparse--one-dense pattern, cannot continuously increase total capacity while keeping active capacity approximately fixed, limiting its pretraining scalability.

\subsection{X-MoD}
\label{sec:xmod_arch}

To remove this bottleneck, we generalize MoD into a sparse-depth hierarchy with independently controllable token sparsity and sparse-depth density.
The resulting architecture, which we call \textbf{X-MoD}, is
\begin{equation}
[Dense]_{\times N_0} + \left([Sparse(K)]_{\times AK} + [Dense]\right)_{\times N_1},
\label{eq:xmod_arch}
\end{equation}
where \(N_0\) denotes an initial dense prefix, \(N_1\) is the number of sparse--dense blocks, and \(A\) is the \emph{anchor stride}, with dense layers serving as anchors.
Intuitively, \(A\) controls how many sparse refinements a token receives on average between two adjacent dense anchors.
Since each sparse layer activates only a \(\frac{1}{K}\) fraction of tokens, placing \(AK\) sparse layers between two dense anchors yields an average of \(AK \cdot \frac{1}{K}=A\) sparse updates per token within one sparse--dense block.
Figure~\ref{fig:XMoD} illustrates the resulting hierarchy.

Let $N_{\mathrm{layer}}$ denote the parameter count of a dense Transformer layer.
Under X-MoD, the total parameter count is
\begin{equation}
N(K,A)=N_{\mathrm{layer}}\bigl(N_0+N_1(1+AK)\bigr),
\label{eq:xmod_total_params}
\end{equation}
while the active-equivalent parameter count, defined as the average number of parameters traversed by a token, is
\begin{equation}
N_{\mathrm{act}}
=
N_{\mathrm{layer}}\bigl(N_0+N_1(1+A)\bigr).
\label{eq:xmod_active_params}
\end{equation}
The total-to-active ratio is therefore \(N(K,A)/N_{\mathrm{act}}\) which grows approximately linearly in \(K\) for fixed \(A\) when the sparse backbone dominates.
This decoupling is the central property that enables X-MoD to explore sparse-depth scaling regimes inaccessible to the original MoD design.

\subsection{Stabilization and token balancing}
\label{sec:xmod_stabilization}

Increasing the total-to-active ratio makes X-MoD more scalable, but stacking many sparse layers between dense anchors can introduce optimization instability and token-routing imbalance. We use three lightweight stabilizers.

\textbf{(i) Layer-wise gating with variance scaling.}\label{sec:layerwise_gating}
Let \(x_i^{\ell}\) denote a selected token at sparse layer \(\ell\).
In the original MoD, the routing score gates only the FFN residual \(f^\ell_i\), while the attention residual \(a^\ell_i\) is injected unscaled. X-MoD instead gates the whole selected-layer residual:
\begin{align}
x_{i,\text{MoD}}^{\ell + 1} &= x_i^{\ell} + a_i^{\ell} + r_i^{\ell} \odot f_i^{\ell}, \\
x_{i,\text{X-MoD}}^{\ell + 1} &= x_i^{\ell} + \varsigma^{\ell} \, r_i^{\ell} \odot (a_i^{\ell}+f_i^{\ell}).
\label{eq:xmod_layer}
\end{align}

This makes the routed computation consistent and stabilizes residual variance through the learnable scale \(\varsigma^{\ell}\).
Figure~\ref{fig:XMoD} compares the original MoD and our X-MoD variant.

\textbf{(ii) Depth-wise token balancing.} \label{sec:depthwise_balancing}
A second failure mode of sparse-depth stacking is \emph{token collapse}: successive sparse layers may repeatedly select the same small subset of tokens.
To discourage this behavior, we propagate a token-choice bias across sparse layers within a sparse--dense block. Let \(s_i^{(\ell)}\) be the router logit of token \(i\) at sparse layer \(\ell\). Before top-\(\frac{1}{K}\) selection, we adjust it as
\begin{equation}
r_i^{(\ell)}
=
\mathrm{sigmoid}\!\left(s_i^{(\ell)}-\tau b_i\right),
\label{eq:token_bias}
\end{equation}
where \(b_i\) increases whenever token \(i\) is selected and is reset at the next dense anchor, discouraging repeated selection without enforcing uniform routing.

\textbf{(iii) Asymmetric hierarchy.} \label{sec:asymmetric_hierarchy}
X-MoD keeps the first \(N_0\) layers dense and introduces sparse-depth routing only in deeper layers.
This reflects the empirical observation that early layers primarily build stable lexical and syntactic representations, whereas deeper layers are more suitable for conditional computation and token-specific refinement.

Taken together, these mechanisms make X-MoD trainable in regimes where the original one-sparse--one-dense MoD architecture becomes too restrictive.

%% file: chapters/ScalingLaw.tex
\begin{figure}[t]
  \centering
  \includegraphics[width=0.485\textwidth]{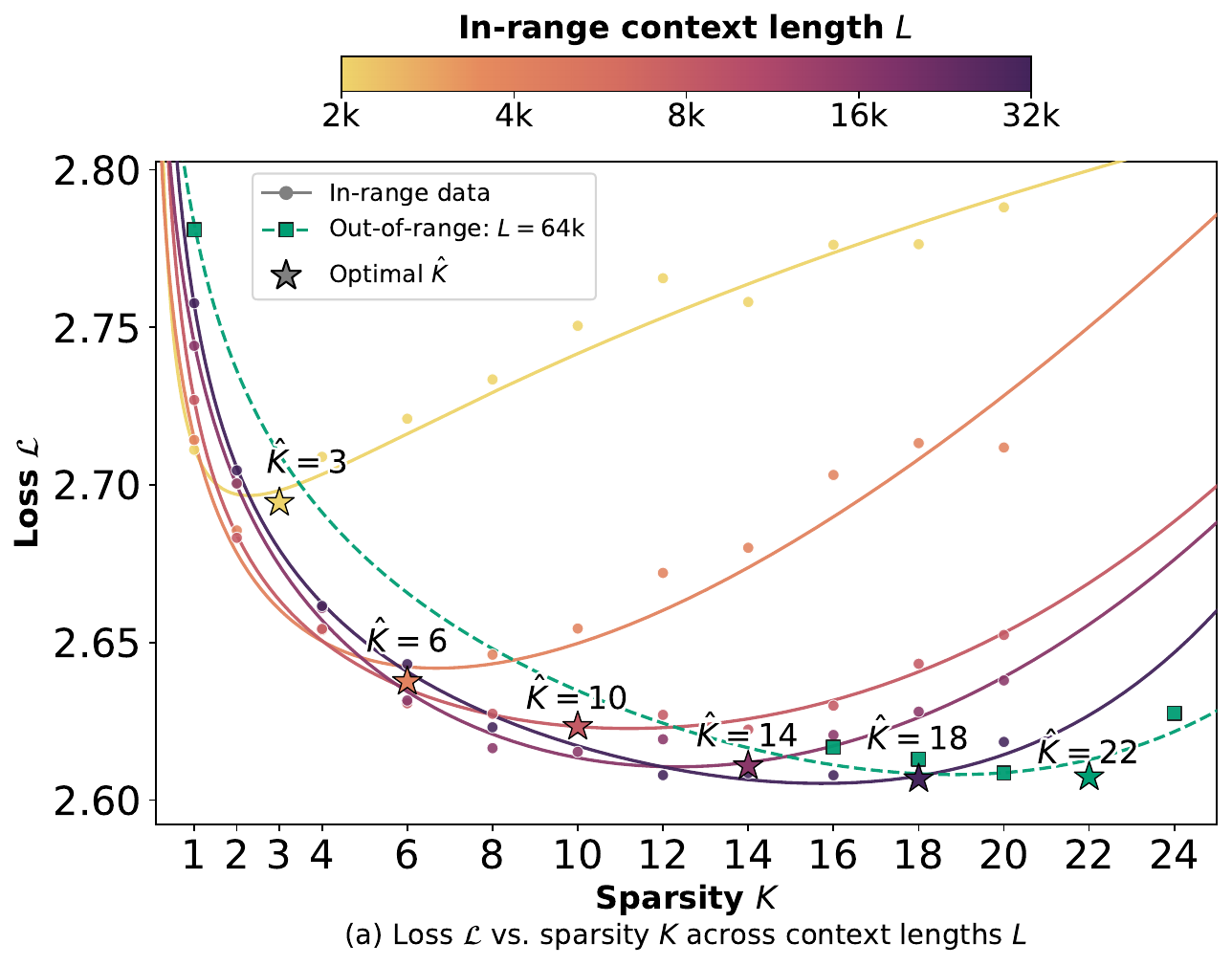}\hfill
  \includegraphics[width=0.485\textwidth]{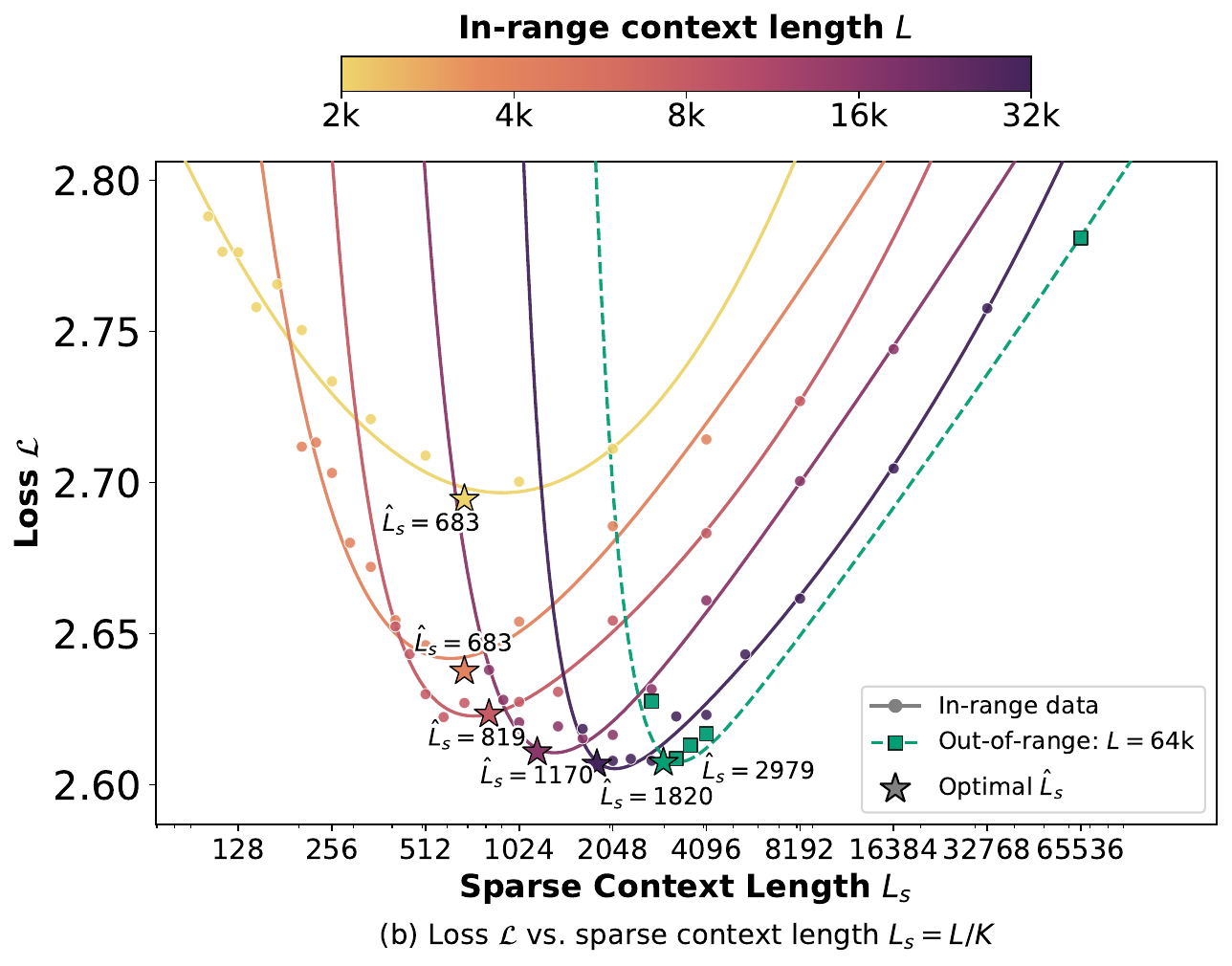}

  \includegraphics[width=0.485\textwidth]{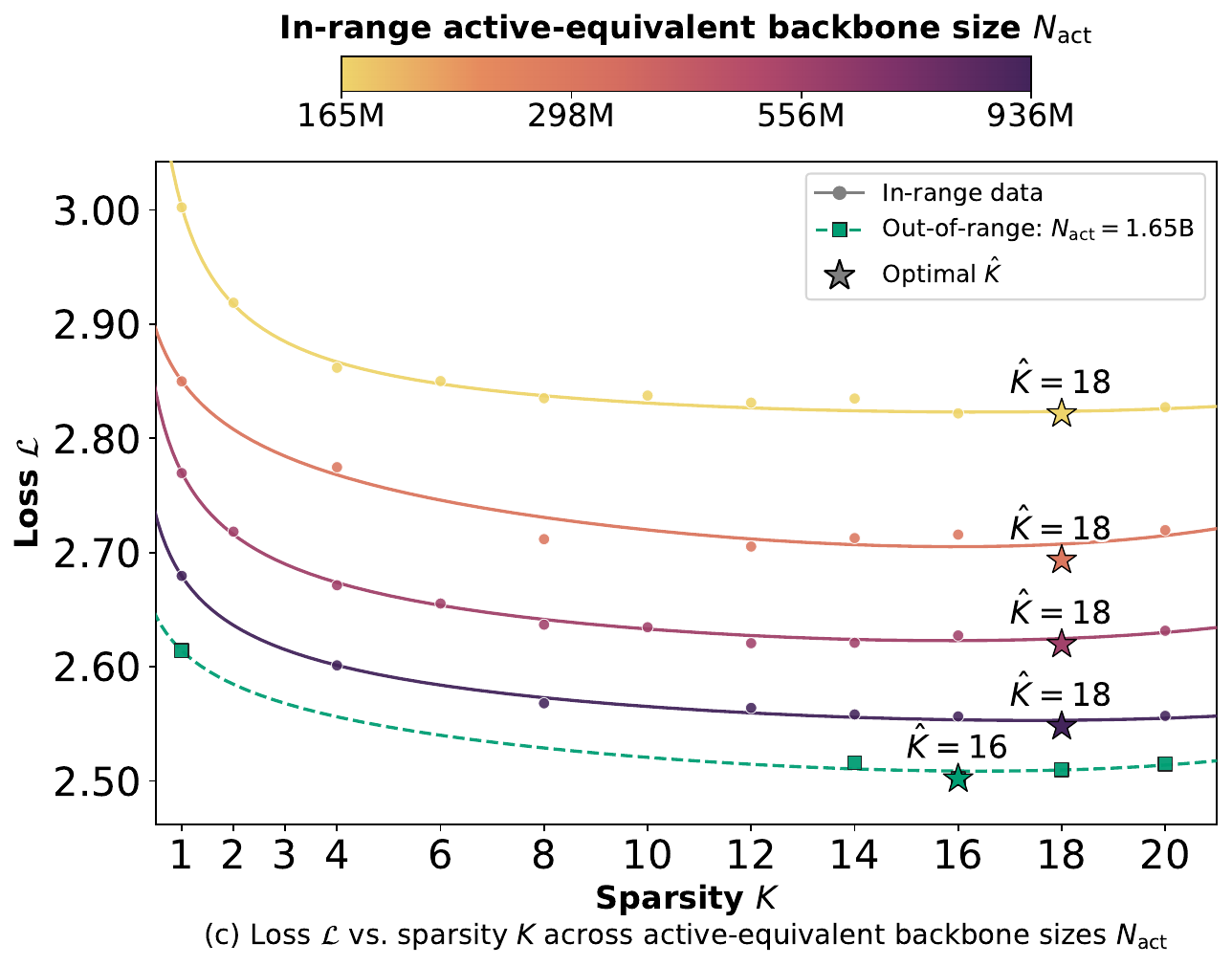}\hfill
  \includegraphics[width=0.485\textwidth]{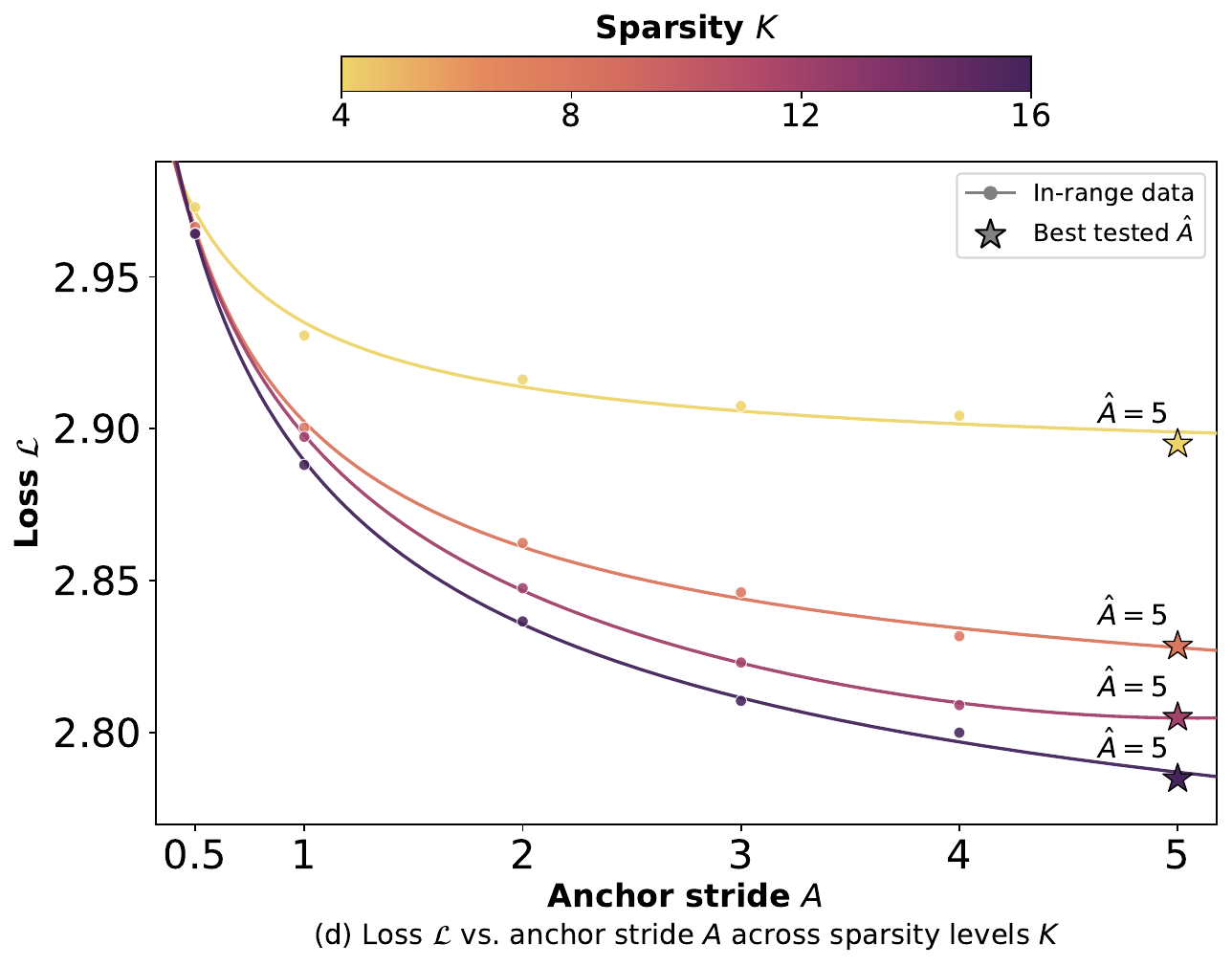}
  \vspace{-6pt}
  \caption{
    \textbf{Empirical regularities motivating the X-MoD scaling law.}
    (a,\,b) Loss is U-shaped in \(K\), and the optimum shifts with context length \(L\), also visible under sparse context \(L_s=L/K\).
    (c) Varying \(N_{\mathrm{act}}\) has weaker effect on optimal \(K\).
    (d) Anchor stride \(A\) interacts with \(K\).
    Sweep settings are given in Appendix~\ref{sec:methods_scaling_sweeps}.
  }
  \label{fig:ScalingLaw}
  \vspace{-12pt}
\end{figure}

\section{Practical Scaling Laws for X-MoD}
\label{sec:xmod_scaling}

\subsection{A conditional design problem}
\label{sec:xmod_design_problem}

The practical goal of X-MoD is to provide a design rule for sparse-depth routing.
Given a training compute budget \(C\), sequence length \(L\), and active-equivalent backbone size \(N_{\mathrm{act}}\), we seek the sparsity \(K\) and anchor stride \(A\) that minimize the final loss:
\begin{equation}
(K^*,A^*)
=
\arg\min_{K,A}
\mathcal L_{\mathrm{X\mbox{-}MoD}}(K,A;C,L,N_{\mathrm{act}}).
\label{eq:xmod_design_objective}
\end{equation}
This is a conditional design problem: \(N_{\mathrm{act}}\) is chosen according to the desired active-capacity or deployment budget, and the scaling law quantifies how routing configurations affect performance within that budget.
To separate sparse-depth effects from the dense backbone, we define a matched-FLOPs residual
\begin{equation}
\Delta(C,L,N_{\mathrm{act}},K,A)
=
\mathcal L_{\mathrm{X\mbox{-}MoD}}(C,L,N_{\mathrm{act}},K,A)
-
\widehat{\mathcal L}_{\mathrm{dense}}(C,L,N_{\mathrm{act}}),
\label{eq:xmod_delta_def}
\end{equation}
where \(\widehat{\mathcal L}_{\mathrm{dense}}\) is obtained by interpolating dense validation curves at matched FLOPs for the same sequence length and backbone family.
This subtracts the contribution of the matched dense reference at the same FLOPs, sequence length and backbone family, leaving the residual variation associated with sparse-depth design to be modeled.

\subsection{Mechanistic constraints and empirical regularities}
\label{sec:xmod_empirical_regularities}

We consider four candidate mechanisms for the residual law.
Increasing total sparse capacity from \(N_{\mathrm{act}}\) to \(N(K,A)\) should reduce loss.
Sparse runs may differ from matched dense baselines in effective token exposure, motivating a candidate exposure correction.
Sparse layers observe only routed sequence subsets, so the law must capture reduced sparse-layer context.
Anchor stride should matter only under active sparse routing, implying an \(A\)-\(K\) interaction.

The sweeps in Fig.~\ref{fig:ScalingLaw} reveal a consistent trade-off.
Loss is U-shaped in \(K\), requiring both a reward for increasing total sparse capacity and a correction for the reduced context observed by each sparse layer.
The optimum \(\hat K\) shifts strongly with context length \(L\), also visible when plotted against sparse context \(L_s=L/K\); by contrast, changing \(N_{\mathrm{act}}\) has a weaker effect.
The \(64\mathrm{k}\) context and \(N_{\mathrm{act}}=1.65\mathrm{B}\) setting extend beyond the ranges used to construct the law and follow the same qualitative trends.
Under fixed-\(N_{\mathrm{act}}\) sweeps, the effect of anchor stride \(A\) also varies with \(K\).
These observations motivate the capacity, context and anchor-stride features of the law.

Inspired by \citet{hoffmann2022} and \citet{abnar2025}, we parameterize the capacity reward and sparse-context correction as
\begin{alignat}{2}
R_N
&=
N_{\mathrm{act}}^{-\eta_N}
&&\left[
1-
\left(\frac{N(K,A)}{N_{\mathrm{act}}}\right)^{-\rho_N}
\right],
\label{eq:xmod_phi_n}\\
P_L
&=
L^{-\eta_L}
&&\left[
\left(\frac{L_s}{L}\right)^{-\rho_L}-1
\right],
\qquad
L_s = L / K.
\label{eq:xmod_phi_l}
\end{alignat}
Here \(N_{\mathrm{act}}\) is evaluated as a parameter count and \(L\) as a token count; their units are absorbed into the fitted coefficients \(u\) and \(w\).
Appendix~\ref{app:candidate_exposure} derives the candidate exposure correction \(P_D\).

For anchor stride, we use the controlled \(A\)-sweep where \(N_{\mathrm{act}}\) is fixed and increasing \(A\) reallocates active-equivalent budget from dense anchors toward sparse refinements.
Appendix~\ref{app:anchor_stride_share} derives the normalized sparse-equivalent share \(\omega(A)=2A/(A+1)\).
Since this benefit appears only when sparse routing is active, we define
\begin{equation}
R_{A,K}
=
\left(K^{\rho_K}-1\right)
\left[
\omega(A)^{\rho_A}-1
\right],
\qquad
\omega(A)=\frac{2A}{A+1}.
\label{eq:xmod_phi_a}
\end{equation}

Each retained feature is tied to a specific sparse-depth mechanism and is anchored to the corresponding dense reference point.

\subsection{A practical scaling law}
\label{sec:xmod_practical_law}

We use the following residual law in the main text:
\begin{equation}
\boxed{
\begin{aligned}
\Delta(C,L,N_{\mathrm{act}},K,A)
&\approx
e
-u\,N_{\mathrm{act}}^{-\eta_N}
\left[
1-
\left(\frac{N(K,A)}{N_{\mathrm{act}}}\right)^{-\rho_N}
\right]\\
&\quad
+w\,L^{-\eta_L}(K^{\rho_L}-1)
-z\,(K^{\rho_K}-1)
\left[
\left(\frac{2A}{A+1}\right)^{\rho_A}-1
\right],
\end{aligned}
}
\label{eq:xmod_final_law}
\end{equation}
where \(u,w,z>0\).
The fitted coefficients and the comparison supporting omission of \(P_D\) are reported in Appendix~\ref{app:xmod_fit_params}.

To study the capacity--context trade-off in isolation, we first fix \(A=1\), for which the anchor-stride interaction vanishes.
For design intuition, using the large-\(K\) approximation \(N(K,A)/N_{\mathrm{act}}\propto K\), Eq.~\eqref{eq:xmod_final_law} reduces up to constants to
\begin{equation}
\Delta_K
\approx
\frac{c_N}{N_{\mathrm{act}}^{\eta_N}K^{\rho_N}}
+
\frac{c_L K^{\rho_L}}{L^{\eta_L}},
\label{eq:xmod_delta_K_asymptotic}
\end{equation}
where the first term decreases with \(K\) and the second increases with \(K\).
Thus \(K\) initially improves performance by increasing total sparse capacity, but excessive sparsity eventually hurts because each sparse layer receives less context.
The same approximation gives the optimal \(K\) value
\begin{equation}
\hat K
\propto
L^{\eta_L/(\rho_N+\rho_L)}
N_{\mathrm{act}}^{-\eta_N/(\rho_N+\rho_L)},
\label{eq:xmod_kstar_scaling}
\end{equation}
with the derivation in Appendix~\ref{app:kstar_derivation}.
This predicts a strong positive dependence on context length and only a weak dependence on active-equivalent model scale, matching the empirical trends in Fig.~\ref{fig:ScalingLaw}.
The \(A\)-term further favors larger sparsity when more active-equivalent budget is allocated to sparse refinements, with saturating gains through the sparse-share ratio \(2A/(A+1)\).

%% file: chapters/Experiments.tex
\section{Experiments}

\subsection{Experimental Settings}
\label{sec:exp_settings}

\textbf{Data and evaluation.}
All models are pretrained on FineWeb-Edu~\citep{finewebedu} tokenized with the GPT-2 tokenizer.
We report validation language-modeling loss on held-out shards.
MoD and X-MoD are trained with non-causal top-\(k\) routing but evaluated with a causal threshold rule~\cite{MoD}; unless otherwise stated, all reported validation losses use this rule.
We also evaluate selected checkpoints on a fixed set of downstream validation tasks: HellaSwag~\citep{benchmark-1-hellaswag}, ARC-Easy and ARC-Challenge~\citep{benchmark-2-arc}, PIQA~\citep{benchmark-3-piqa}, LAMBADA~\citep{benchmark-4-lambda} and BoolQ~\citep{benchmark-5-boolq}.

\textbf{Training and FLOPs accounting.}
All models are trained under the same recipe, with a fixed token budget per optimization step across sequence lengths \(L\).
Training curves and model comparisons are aligned by logged training FLOPs.
The training recipe, model configurations, compute budgets and evaluation protocols are collected in Appendix~\ref{app:training_details}.

\textbf{Scaling-law sweeps.}
We vary sparsity \(K\), context length \(L\), active-equivalent size \(N_{\mathrm{act}}\) and anchor stride \(A\), comparing each sparse configuration with its dense reference at matched training compute (Fig.~\ref{fig:ScalingLaw}).
The \(L=64\mathrm{k}\) and \(N_{\mathrm{act}}=1.65\mathrm{B}\) settings are prespecified out-of-range extensions and are excluded from fitting the multivariable law.
Sweep grids and fitting checkpoints are specified in Appendix~\ref{sec:methods_scaling_sweeps}.

\subsection{Out-of-Sample Validation}
\label{sec:oos_validation}

A useful design law should predict held-out settings, not merely interpolate the sweeps used to fit it.
We evaluate Eq.~\eqref{eq:xmod_final_law} by withholding complete context-length, anchor-stride or model-scale groups, refitting on the remaining observations, and predicting validation loss for the excluded configurations (Appendix~\ref{app:oos_protocol}).

\begin{figure}[t]
  \centering
  \includegraphics[width=\textwidth]{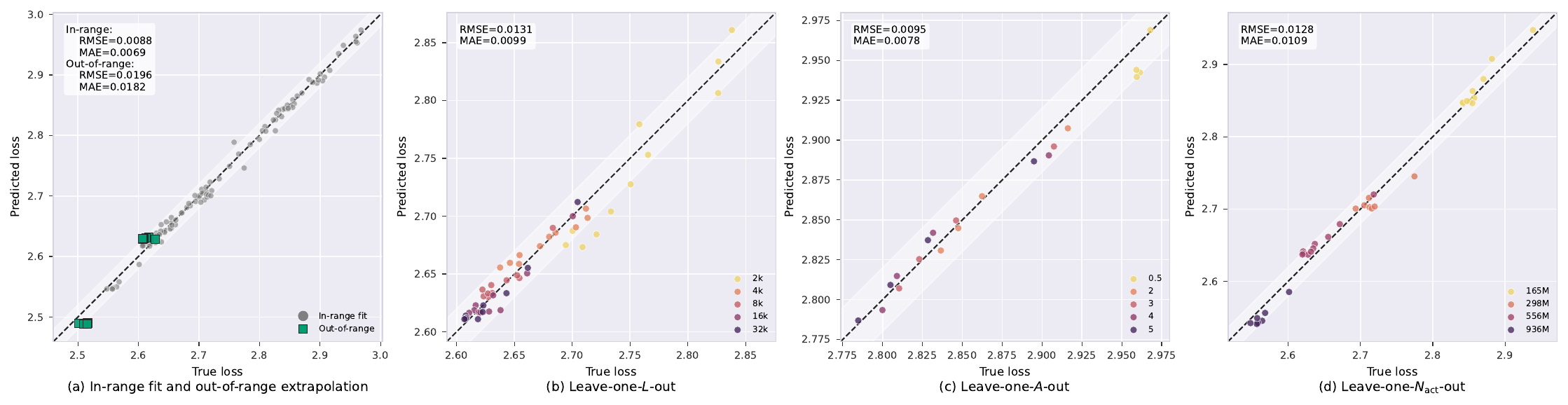}
  \vspace{-12pt}
  \caption{
    \textbf{Predicted vs. true validation loss.}
    (a) In-range fit (circles) and out-of-range predictions without refitting (squares).
    (b--d) Leave-one-group-out predictions by \(L\), \(A\) and \(N_{\mathrm{act}}\), respectively.
  }
  \label{fig:scaling_validation}
  \label{fig:LOO}
  \vspace{-12pt}
\end{figure}

Figure~\ref{fig:scaling_validation}a shows close agreement between predictions and measurements, both within the fitting range and at the longer-context and larger-model extensions.
The grouped tests (Fig.~\ref{fig:scaling_validation}b--d) retain this accuracy, indicating that the law captures transferable structure in sparse-depth routing.
Fitted parameters and in-range goodness of fit are reported in Appendix~\ref{app:xmod_fit_params}.

\subsection{Main Results against Dense, MoD, and MoE}
\label{sec:main_results}

MoE is the closest sparse-capacity analogue to X-MoD: both decouple total capacity from active computation, but MoE routes across width while X-MoD routes across depth.
We first grid-search MoE sparsity and expert granularity at \(N_{\mathrm{act}}=165\mathrm{M}\) under a matched training-compute budget.
The search selects MoE 8:64 and 8:128 as the lowest-loss baselines in their respective sparsity groups, using the established fine-grained and shared-expert design~\citep{DeepSeekMoE2024}.

For each MoE baseline, we then select the lowest-loss X-MoD configuration in the pilot grid with the same \(N_{\mathrm{act}}\) and nominal sparsity \(K\), without exceeding its total-parameter budget.
This yields X-MoD A4K8 and A3K16, respectively (Appendix~\ref{sec:methods_pilot_selection} and Fig.~\ref{fig:xmod_moe_pilot_grid}).
At larger scales, we retain these settings and match \(N_{\mathrm{act}}\) and \(C\) across models.

Across the evaluated scales (Table~\ref{tab:main_results_compact}), X-MoD consistently improves over Dense and the original MoD, and is competitive with or better than representative MoE baselines at similar total parameter scale and training compute, while using substantially lower per-token FLOPs.
The downstream averages generally follow the validation-loss trend, suggesting that the gains are not limited to the pretraining metric.
Measured systems performance at the 936M active-equivalent scale further shows higher training and forward-inference throughput than the approximately total-parameter-matched MoE baselines, at comparable peak memory (Appendix Table~\ref{tab:systems_benchmark}).
Together, these results support sparse depth as an effective alternative to sparse width for expanding model capacity.

\begin{table}[t]
\centering
\caption{
\textbf{Main results under matched active-equivalent size and training compute.}
\(N\) denotes total parameters, and \(\phi/\phi_D\) denotes per-token training FLOPs normalized by the dense model with the same \(N_{\mathrm{act}}\) and sequence length.
The best results are highlighted in \textbf{bold}.
Detailed compute budgets and full model configurations are reported in Appendix~\ref{app:training_details}.
}
\label{tab:main_results_compact}
\label{tab:full_task_results}
\vspace{3pt}
\scriptsize
\setlength{\tabcolsep}{3.2pt}
\begin{tabular}{l l r r r r r r r r r r}
\toprule
\(N_{\mathrm{act}}\) & Model & \(\qquad N\) & \(\phi/\phi_D\downarrow\) & Loss \(\downarrow\) & Hella. \(\uparrow\) & ARC-e \(\uparrow\) & ARC-c \(\uparrow\) & PIQA \(\uparrow\) & LAMB. \(\uparrow\) & BoolQ \(\uparrow\) & Avg. \(\uparrow\) \\
\midrule
\multirow{6}{*}{556M}
& Dense         & 0.54B & 1.00 & 2.773 & 30.42 & 51.93 & 23.21 & 63.80 & 18.47 & 60.52 & 41.39 \\
& MoD           & 0.87B & 0.92 & 2.692 & 34.00 & 58.75 & 25.94 & 66.43 & 21.33 & 51.99 & 43.07 \\
& MoE 8:64      & 2.46B & 1.00 & 2.564 & 36.62 & 62.25 & 27.30 & 68.99 & 26.37 & \textbf{60.95} & 47.08 \\
& X-MoD A4K8    & 2.64B & 0.47 & 2.549 & 38.30 & 62.08 & 27.56 & 68.34 & 29.19 & 58.62 & 47.35 \\
& MoE 8:128     & 4.58B & 1.00 & 2.528 & \textbf{38.57} & \textbf{64.98} & \textbf{30.38} & \textbf{70.18} & 27.25 & 56.30 & 47.94 \\
& X-MoD A3K16   & 4.79B & \textbf{0.46} & \textbf{2.503} & 38.00 & 64.31 & 28.41 & 70.13 & \textbf{30.49} & 60.49 & \textbf{48.64} \\
\midrule
\multirow{6}{*}{936M}
& Dense         & 936M  & 1.00 & 2.635 & 35.18 & 59.85 & 26.71 & 67.52 & 24.16 & 59.85 & 45.54 \\
& MoD           & 1.48B & 0.92 & 2.570 & 36.67 & 62.04 & 26.54 & 68.72 & 26.14 & \textbf{60.89} & 46.83 \\
& MoE 8:64      & 4.51B & 1.00 & 2.445 & 40.62 & 67.76 & 32.34 & 71.22 & 33.11 & 60.80 & 50.97 \\
& X-MoD A4K8    & 4.52B & 0.51 & 2.427 & 41.26 & 68.69 & 33.45 & 71.22 & 33.55 & 60.43 & 51.43 \\
& MoE 8:128     & 8.48B & 1.00 & 2.412 & 41.30 & \textbf{70.16} & \textbf{33.96} & 70.78 & 34.23 & 60.06 & 51.75 \\
& X-MoD A3K16   & 8.24B & \textbf{0.50} & \textbf{2.402} & \textbf{42.07} & 67.80 & 33.79 & \textbf{72.03} & \textbf{34.64} & 60.67 & \textbf{51.84} \\
\midrule
\multirow{6}{*}{1.65B}
& Dense         & 1.65B & 1.00 & 2.531 & 37.66 & 63.30 & 28.50 & 69.64 & 28.45 & 61.22 & 48.13 \\
& MoD           & 2.67B & 0.93 & 2.502 & 40.69 & 65.36 & 29.10 & 70.08 & 29.46 & 60.89 & 49.26 \\
& MoE 8:64      & 9.28B & 1.00 & 2.367 & 42.43 & 67.51 & 33.79 & 72.31 & 35.28 & 61.41 & 52.12 \\
& X-MoD A4K8    & 8.31B & 0.57 & 2.359 & 42.99 & 69.02 & 33.19 & 72.03 & 34.45 & \textbf{62.02} & 52.28 \\
& MoE 8:128     & 17.16B & 1.00 & 2.333 & \textbf{43.56} & 70.66 & \textbf{35.32} & \textbf{72.96} & 35.47 & 61.62 & 53.27 \\
& X-MoD A3K16   & 14.73B & \textbf{0.56} & \textbf{2.322} & 43.47 & \textbf{71.25} & \textbf{35.32} & 72.58 & \textbf{37.45} & 61.04 & \textbf{53.52} \\
\bottomrule
\end{tabular}
\vspace{-12pt}
\end{table}

\subsection{Ablations}
\label{sec:ablations}
\label{sec:results_ablations}

Tables~\ref{tab:xmod_mechanism_controls} and \ref{tab:xmod_component_ablations} examine the sources of X-MoD's gains through mechanism controls and component removals, respectively.
Detailed protocols are provided in Appendix~\ref{sec:methods_ablations}.

\begin{table}[t]
\centering
\caption{
\textbf{Mechanism controls of X-MoD.}
All variants use the 936M reference backbone.
Metrics follow Table~\ref{tab:main_results_compact}; configurations are given in Appendix~\ref{sec:methods_ablations}.
}
\label{tab:xmod_mechanism_controls}
\vspace{3pt}
\scriptsize
\setlength{\tabcolsep}{3.2pt}
\begin{tabular}{l r r r r r r r r r r}
\toprule
Variant & Total \(N\) & \(\phi/\phi_D\) & Loss \(\downarrow\) & Hella. \(\uparrow\) & ARC-e \(\uparrow\) & ARC-c \(\uparrow\) & PIQA \(\uparrow\) & LAMB. \(\uparrow\) & BoolQ \(\uparrow\) & Avg. \(\uparrow\) \\
\midrule
Dense & 936M & 1.00 & 2.635 & 35.18 & 59.85 & 26.71 & 67.52 & 24.16 & 59.85 & 45.54 \\
MoD & 1.48B & 0.92 & 2.570 & 36.67 & 62.04 & 26.54 & 68.72 & 26.14 & 60.89 & 46.83 \\
\midrule
Routed-FFN A4K8 & 3.26B & 1.00 & 2.508 & 37.97 & 64.27 & 28.92 & 70.29 & 29.79 & 60.95 & 48.70 \\
Routed-FFN A3K16 & 5.68B & 1.00 & 2.482 & 40.71 & 65.61 & 30.38 & 69.15 & 29.89 & 60.37 & 49.35 \\
\midrule
Full X-MoD A4K8 & 4.52B & 0.51 & \textbf{2.427} & 41.26 & 68.69 & 33.45 & 71.22 & 33.55 & 60.43 & \textbf{51.43} \\
Selected-Q / full-KV A4K8 & 4.52B & 1.09 & 2.564 & 38.44 & 61.66 & 27.65 & 68.01 & 28.90 & 57.71 & 47.06 \\
Full X-MoD A3K16 & 8.24B & 0.50 & \textbf{2.402} & 42.07 & 67.80 & 33.79 & 72.03 & 34.64 & 60.67 & \textbf{51.84} \\
Selected-Q / full-KV A3K16 & 8.24B & 1.19 & 2.558 & 38.35 & 62.04 & 27.56 & 68.28 & 29.13 & 58.62 & 47.33 \\
\bottomrule
\vspace{-18pt}
\end{tabular}
\end{table}

\begin{table}[t]
\centering
\caption{
\textbf{Ablations of X-MoD stabilization mechanisms.}
All variants are compared under the same \(N_{\mathrm{act}}=936\)M.
\(\Delta\)Loss is measured relative to Full X-MoD; task scores follow Table~\ref{tab:main_results_compact}.
}
\label{tab:xmod_component_ablations}
\label{tab:xmod_ablation}
\label{tab:xmod_ablations}
\vspace{3pt}
\scriptsize
\setlength{\tabcolsep}{3.2pt}
\begin{tabular}{l r r r r r r r r r}
\toprule
Variant & Loss \(\downarrow\) & \(\Delta\)Loss & Hella. \(\uparrow\) & ARC-e \(\uparrow\) & ARC-c \(\uparrow\) & PIQA \(\uparrow\) & LAMB. \(\uparrow\) & BoolQ \(\uparrow\) & Avg. \(\uparrow\) \\
\midrule
Dense reference & 2.635 & +0.159 & 35.18 & 59.85 & 26.71 & 67.52 & 24.16 & 59.85 & 45.54 \\
Full X-MoD A1K16 & \textbf{2.476} & 0.000 & 40.64 & 66.08 & 30.80 & 71.27 & 30.58 & 61.31 & \textbf{50.12} \\
\quad w/o dense anchors & 2.579 & +0.103 & 36.35 & 62.63 & 28.33 & 69.53 & 26.59 & 58.96 & 47.06 \\
\quad w/o gated residual scaling & 2.492 & +0.016 & 40.62 & 65.19 & 29.27 & 69.31 & 29.73 & 60.58 & 49.12 \\
\quad w/o token-choice bias & 2.657 & +0.181 & 34.59 & 59.85 & 26.54 & 67.52 & 22.01 & 59.11 & 44.94 \\
\quad w/o dense prefix & 2.487 & +0.011 & 40.41 & 65.19 & 30.20 & 69.70 & 29.69 & 61.01 & 49.37 \\
\bottomrule
\vspace{-18pt}
\end{tabular}
\end{table}

\textbf{Conditional capacity.}
We first use two counterfactual controls at the 936M reference scale to separate conditional-depth capacity from sparse-layer attention-context selection (Table~\ref{tab:xmod_mechanism_controls}).
The Routed-FFN control repeats \(A\) groups of full-sequence attention followed by \(K\) routed FFN refinements between dense anchors.
Both Routed-FFN variants improve on Dense at the same active-equivalent size, per-token FLOPs and training compute, with further gains from the larger-capacity variant.
This supports a contribution from increased conditional parameter capacity even without sparse-layer attention-context selection.

\textbf{Attention context.}
The Selected-Q/full-KV diagnostic in Table~\ref{tab:xmod_mechanism_controls} instead retains the router, selected queries, total depth and total parameters of X-MoD, while allowing all tokens to provide keys and values.
The additional K/V projections increase active computation and per-token FLOPs.
Under this fixed-\(C\) objective, the full X-MoD design offers a more effective allocation of computation than expanding the sparse-layer KV context.
Detailed control configurations are given in Appendix Table~\ref{tab:counterfactual_configurations}.

\textbf{Stabilization mechanisms.}
We then ablate the stabilization mechanisms introduced in Section~\ref{sec:xmod_stabilization} on the 936M model.
All variants use the same \(N_{\mathrm{act}}\), context length, and training compute, so the comparison isolates the effect of each X-MoD component.
Table~\ref{tab:xmod_component_ablations} shows that each mechanism contributes to sparse-depth training.
Removing dense anchors causes a large degradation, indicating that periodic full-token synchronization is important when many sparse layers are stacked.
Removing depth-wise token-choice bias causes the largest degradation, confirming token concentration as a major failure mode in sparse-depth routing.
Gated residual scaling gives a smaller but consistent improvement by stabilizing the selected-layer residual, while the dense prefix improves early representation learning.
Removing any of the four components increases validation loss and lowers the downstream average relative to the full model.

\textbf{Token selection.}
Routing diagnostics directly probe repeated token selection across depth (Appendix Table~\ref{tab:routing_diagnostics}).
For A1K16, removing token-choice bias causes routing to collapse onto a persistent token subset: consecutive sparse layers repeatedly select highly overlapping token sets, with a pronounced concentration of tokens updated in all 16 sparse layers (Appendix Fig.~\ref{fig:routing_distributions}a).
The full A4K8 and A3K16 models combine near-complete interval coverage with low consecutive-layer overlap (Appendix Table~\ref{tab:routing_diagnostics} and Fig.~\ref{fig:routing_distributions}b,c).
Across all three full-model configurations, threshold and top-\(k\) masks show high agreement when computed from the same routing scores, supporting close alignment between the training-time and deployable selection rules.

\textbf{Depth scaling.}
Stable training through 412 total layers demonstrates that the architectural changes support substantial sparse-depth expansion (Appendix Table~\ref{tab:xmod_depth_scaling}).

%% file: chapters/Limitations.tex
\section{Limitations and Future Work}
\label{sec:limitations}

This work studies X-MoD as a sparse-depth architecture in isolation, but practical sparse models often combine multiple forms of conditional computation.
For example, X-MoD could potentially be combined with sparse-width routing such as MoE, recurrent-depth mechanisms such as Loop Transformers, or inference-time token pruning~\cite{MoE2017,mor-1-mor,mor-4-loopformer,sparse-attn-1-mixture-of-sparse-attention,sparse-attn-5-nsa,skiplayer-1-layerskip,skiplayer-3-not-all-layers}.
A central open question is whether the scaling laws of these mechanisms compose additively or interact nonlinearly.
This is especially important because sparse activation along depth may weaken gradient stability when many conditional layers are stacked, while MoE introduces its own routing imbalance and expert-specialization dynamics.
Understanding the joint scaling behavior of sparse depth, sparse width, and recurrent computation is therefore an important direction for future work.

The fitted law is evaluated across the model sizes, context lengths, sparsity levels, and anchor strides studied here, including held-out groups and the out-of-range configurations in Figure~\ref{fig:scaling_validation}.
Extending these tests to larger models, longer training horizons, additional seeds, and broader compute budgets would establish how the design rules transfer to further pretraining regimes.
In particular, future work should test whether the predicted dependence of optimal sparsity on context length and active-equivalent scale continues to hold at production-scale model sizes, and whether the anchor-stride gains remain stable when sparse stacks become substantially deeper.

Our systems measurements quantify the throughput gains of X-MoD in the evaluated implementation (Table~\ref{tab:systems_benchmark}).
X-MoD also introduces opportunities for further optimization: consecutive sparse layers may process partially disjoint token subsets, dense anchors provide natural synchronization points, and routing decisions can potentially be reused or pipelined across depth.
Developing efficient kernels, scheduling strategies, and parallelization schemes for sparse-depth routing is therefore an important direction for making X-MoD practical at larger scales.

%% file: chapters/RelatedWorks.tex
\section{Related works}
\label{sec:related_work}

\paragraph{Sparse conditional computation.}
Mixture-of-Experts (MoE) models decouple total parameter count from per-example computation by routing each token to a subset of feed-forward experts, enabling substantial capacity growth under a nearly fixed active budget~\citep{MoE2017,GShard2020,SwitchTransformer2022,DeepSeekMoE2024}. Mixture-of-Depths (MoD) applies a related idea along the depth dimension: only selected tokens pass through certain Transformer layers, while unselected tokens bypass them~\citep{MoD}. 
Several recent works explore MoD-like token routing in adjacent settings. In multimodal models, conditional depth is often applied to image or vision tokens, whose redundancy makes token-level sparsification especially attractive~\citep{mllm-1-moma,mllm-2-mole-vla,mllm-3-videollm-mod,mllm-8-gamma-mod,mllm-9-p-mod}. 
Another line of work uses token skipping or layer pruning during fine-tuning or inference to reduce downstream compute while preserving task performance~\citep{skiplayer-1-layerskip,skiplayer-2-shortened-llama,skiplayer-3-not-all-layers,skiplayer-4-d-llm}. These directions primarily target multimodal efficiency or downstream budget reduction, whereas our focus is pretraining-time sparse-depth scalability and design rules under fixed compute. 
Other sparse-compute mechanisms, including sparse attention~\citep{sparse-attn-1-mixture-of-sparse-attention,sparse-attn-2-streamingllm,sparse-attn-3-h2o,sparse-attn-4-model-tells,sparse-attn-5-nsa}, recurrent or looped depth models~\citep{mor-1-mor,mor-2-inner-thinking-transformer,mor-4-loopformer,mor-5-adaptive-loops,mor-6-less-is-more,mor-7-hierarchical-reasoning-model}, and null-expert token skipping~\citep{nullexpert-1-adamoe,nullexpert-2-moe++,nullexpert-3-longcat}, are complementary to our setting: they reduce or reuse computation rather than directly studying how to expand total sparse-depth capacity at fixed active-equivalent budget.

\paragraph{Scaling laws.}
Scaling laws for language models show that model size, data, and compute interact predictably in dense Transformers~\citep{kaplan2020,hoffmann2022}. Subsequent work extends this perspective to other axes such as context length, data mixtures, and sparsely activated architectures~\citep{xiong2024effective,team2024gemini}. In particular, recent MoE scaling-law studies separate total parameters from active parameters and analyze how sparsity affects compute-optimal model design~\citep{clark2022,GLaM2022,ludziejewski2024,wang2024,abnar2025,jointmoescalinglaws2025,tian2025}. 
In contrast, sparse-depth scaling remains underexplored. 

%% file: chapters/Conclusion.tex
\section{Conclusion}
\label{sec:conclusion}

We introduced X-MoD, a scalable sparse-depth architecture that extends Mixture-of-Depths beyond the original one-sparse--one-dense regime. 
By decoupling token sparsity from anchor stride, X-MoD can increase total capacity while keeping active-equivalent capacity controlled. 
We further formulated sparse-depth routing as a conditional architecture-design problem and developed a practical scaling law over a FLOP-matched dense baseline. 
The resulting law decomposes X-MoD behavior into sparse-capacity reward, sparse-context correction, and anchor-stride interaction, providing quantitative guidance for sparse-depth configuration. 
Empirically, X-MoD configurations improve over Dense and MoD, remain competitive with representative MoE baselines, and are supported by held-out loss prediction and ablations. 
Overall, our results suggest that sparse-depth routing is a viable and quantitatively analyzable axis for scaling conditional-computation Transformers.

%% file: chapters/Acknowledgments.tex
\section*{Acknowledgments}
This work was supported in part by the National Natural Science Foundation of China under Grant Nos. 62676210, 62272264, and the National Key Research and Development Program of China under Grant No. 2020YFA0804503.

%% file: chapters/AIUse.tex
\section*{Disclosure of generative AI use}
In this work, we used generative AI tools to assist with translation. We did not use generative AI tools to help develop theoretical models or conceptual frameworks, formulate mathematical claims, provide key elements for proving mathematical claims, assist in writing proofs, propose or refine hypotheses, design or provide feedback on research methods or experiments, implement methods, support qualitative and thematic data analysis, or interpret results. Generating synthetic datasets and cleaning or reformatting datasets are not applicable to this work. Additionally, we used generative AI tools to edit the paper for readability. We have reviewed all AI-assisted work. We take responsibility for the final content of this work, including text, claims, or artifacts produced with the assistance of generative AI.

%% file: chapters/Appendix.tex
\section{Scaling-law derivations}
\label{app:derivations}

\subsection{Anchor-stride sparse-equivalent share}
\label{app:anchor_stride_share}

We derive the normalized sparse-equivalent share used in the anchor-stride term. Consider the controlled \(A\)-sweep in which the active-equivalent parameter budget is held fixed. Starting from the X-MoD stage
\begin{equation}
[Dense]_{\times N_0}
+
\left([Sparse(K)]_{\times AK}+[Dense]\right)_{\times N_1},
\end{equation}
changing \(A\) while keeping the active-equivalent budget fixed is implemented by changing the number of stages. Let
\begin{equation}
H=\frac{N_{\mathrm{act}}}{N_{\mathrm{layer}}}-N_0
\end{equation}
denote the active-equivalent depth allocated to the non-prefix part of the model. Since each stage contributes \(A\) sparse updates per token and one dense-anchor update, the number of stages under anchor stride \(A\) is
\begin{equation}
N_1(A)=\frac{H}{A+1}.
\end{equation}
The sparse part of the active-equivalent depth is
\begin{equation}
\ell_{\mathrm{sp}}(A)
=
A\,N_1(A)
=
\frac{A}{A+1}H,
\end{equation}
and the dense-anchor part is
\begin{equation}
\ell_{\mathrm{anc}}(A)
=
N_1(A)
=
\frac{1}{A+1}H.
\end{equation}
Their sum is \(H\), which is independent of \(A\). Thus, in this controlled sweep, changing \(A\) reallocates the non-prefix active path between sparse refinements and dense anchors without changing the total active-equivalent budget.

The sparse-equivalent share is therefore
\begin{equation}
s(A)
=
\frac{\ell_{\mathrm{sp}}(A)}
{\ell_{\mathrm{sp}}(A)+\ell_{\mathrm{anc}}(A)}
=
\frac{A}{A+1}.
\end{equation}
Normalizing by the \(A=1\) baseline gives
\begin{equation}
\omega(A)
=
\frac{s(A)}{s(1)}
=
\frac{2A}{A+1}.
\end{equation}
This quantity is monotone and saturating in \(A\), equals \(1\) at \(A=1\), and approaches \(2\) as \(A\to\infty\). We use \(\omega(A)\) because it captures the fraction of active-equivalent computation allocated to sparse refinements, rather than treating raw \(A\) as an uninterpreted hyperparameter.

\subsection{Candidate exposure correction and compute-to-token relation}
\label{app:candidate_exposure}
\label{app:compute_token_relation}
Following prior scaling-law studies \citep{hoffmann2022,abnar2025}, we first evaluate the full candidate form
\begin{equation}
\Delta \approx e-uR_N+vP_D+wP_L-zR_{A,K},
\qquad u,v,w,z\geq0,
\label{eq:xmod_general_residual}
\end{equation}
with
\begin{equation}
P_D =
D_{\mathrm{dense}}^{-\eta_D}
\left[
\left(\frac{D_{\mathrm{sparse}}/K}{D_{\mathrm{dense}}}\right)^{-\rho_D}-1
\right].
\label{eq:xmod_phi_d}
\end{equation}
The candidate \(P_D\) term tests whether this exposure difference contributes information beyond the capacity and context terms.

\textbf{Compute-to-token budget relation.}
We derive the relation between the matched compute budget \(C\) and the token budgets \(D_{\mathrm{sparse}}\) and \(D_{\mathrm{dense}}\) using a simplified per-token FLOPs model. Let \(\lambda_p\) denote the constant multiplying token-linear projection and FFN terms, and let \(\lambda_m\) denote the constant multiplying attention matrix-multiplication terms. For an X-MoD model with sequence length \(L\), width \(d\), sparsity \(K\), and anchor stride \(A\), the per-token FLOPs can be written as
\begin{equation}
\phi_{\mathrm{X\mbox{-}MoD}}(L,K,A)
=
\lambda_p d^2\bigl[N_0+N_1(1+A)\bigr]
+
\lambda_m Ld\left[N_0+N_1\left(1+\frac{A}{K}\right)\right].
\label{eq:appendix_xmod_pertoken_flops}
\end{equation}
The first term counts token-linear projections and FFN computation. Since each of the \(AK\) sparse layers processes only a \(1/K\) fraction of tokens, these sparse layers contribute an average of \(A\) active layers per token. The second term counts attention matrix multiplication. Dense prefix and anchor layers attend over the full sequence, while sparse layers attend over routed subsets, yielding the \(N_1A/K\) contribution.

For the matched dense baseline with the same active-equivalent depth \(N_0+N_1(1+A)\), the per-token FLOPs are
\begin{equation}
\phi_{\mathrm{dense}}(L,A)
=
\lambda_p d^2\bigl[N_0+N_1(1+A)\bigr]
+
\lambda_m Ld\bigl[N_0+N_1(1+A)\bigr].
\label{eq:appendix_dense_pertoken_flops}
\end{equation}
Thus, under a fixed training compute budget \(C\),
\begin{equation}
D_{\mathrm{sparse}}(C,L,K,A)
=
\frac{C}{\phi_{\mathrm{X\mbox{-}MoD}}(L,K,A)},
\qquad
D_{\mathrm{dense}}(C,L,A)
=
\frac{C}{\phi_{\mathrm{dense}}(L,A)}.
\label{eq:appendix_token_budget}
\end{equation}
The effective exposure of sparse parameters is proportional to \(D_{\mathrm{sparse}}/K\), since each sparse layer observes only a \(1/K\) fraction of tokens. Therefore the exposure ratio used in the candidate correction can be motivated as
\begin{equation}
\frac{D_{\mathrm{sparse}}/K}{D_{\mathrm{dense}}}
=
\frac{1}{K}
\frac{\phi_{\mathrm{dense}}(L,A)}{\phi_{\mathrm{X\mbox{-}MoD}}(L,K,A)}.
\label{eq:appendix_exposure_ratio}
\end{equation}
This analytic form is used only to motivate the candidate exposure correction. In all empirical fits, we compute \(D_{\mathrm{sparse}}\), \(D_{\mathrm{dense}}\), and \(C\) from the actual logged token counts and FLOPs, which avoids relying on implementation-specific FLOPs approximations.

\subsection{Approximate optimum sparsity}
\label{app:kstar_derivation}

We first fix \(A=1\) to isolate the capacity--context trade-off. The anchor-stride interaction then vanishes, giving
\begin{equation}
\Delta
\approx
e
-
uN_{\mathrm{act}}^{-\eta_N}
\left[
1-
\left(\frac{N(K,A)}{N_{\mathrm{act}}}\right)^{-\rho_N}
\right]
+
wL^{-\eta_L}(K^{\rho_L}-1).
\end{equation}
In the large-\(K\) regime where \(N(K,A)/N_{\mathrm{act}}\propto K\), constants can be absorbed into \(c_N,c_L>0\), giving
\begin{equation}
\Delta_K
\approx
\frac{c_N}{N_{\mathrm{act}}^{\eta_N}K^{\rho_N}}
+
\frac{c_LK^{\rho_L}}{L^{\eta_L}}.
\end{equation}
Setting \(\partial \Delta_K/\partial K=0\) yields
\begin{equation}
-\frac{c_N\rho_N}{N_{\mathrm{act}}^{\eta_N}}K^{-\rho_N-1}
+
\frac{c_L\rho_L}{L^{\eta_L}}K^{\rho_L-1}
=0.
\label{eq:xmod_kstar_stationarity}
\end{equation}
Therefore
\begin{equation}
K^{\rho_N+\rho_L}
=
\frac{c_N\rho_N}{c_L\rho_L}
\frac{L^{\eta_L}}{N_{\mathrm{act}}^{\eta_N}},
\label{eq:xmod_kstar_power_relation}
\end{equation}
and
\begin{equation}
K^*
\propto
L^{\eta_L/(\rho_N+\rho_L)}
N_{\mathrm{act}}^{-\eta_N/(\rho_N+\rho_L)}.
\end{equation}
This expression is intended as an asymptotic design intuition; the representative configurations in the main comparison are selected using the matched-budget pilot grid in Fig.~\ref{fig:xmod_moe_pilot_grid}.

\section{Experimental settings and protocols}
\label{app:training_details}

\subsection{Training and evaluation}

\textbf{Training recipe.}
All models use a warmup-then-cosine learning-rate schedule, with \(1\%\) warmup and a minimum learning rate equal to \(10\%\) of the peak learning rate. Across sequence lengths, we keep the number of tokens per optimization step approximately fixed:
\begin{equation}
\texttt{global\_batch\_size}\times L \approx 2.5\mathrm{M}.
\end{equation}

\textbf{Model details.}
All models use GQA attention~\cite{GQA}. Dense, MoD, and X-MoD use standard feed-forward layers, while MoE baselines replace the feed-forward module with routed experts and shared experts. The 556M and 1.65B active-equivalent backbones follow the architectural configurations of Qwen3-0.6B and Qwen3-1.7B, respectively~\cite{qwen3technicalreport}; other sizes are obtained by varying only hidden dimension, head dimension, and FFN intermediate size. Table~\ref{tab:backbone_configs} summarizes the backbone dimensions, including the 165M and 298M backbones used in the scaling-law sweeps; the GPT-2 vocabulary contains 50,257 tokens. Table~\ref{tab:main_model_configurations} reports the full main-comparison configurations, including physical layer count, expert configuration, total parameters, FLOPs per token and execution mode.

\textbf{MoE routing.}
Drawing on the load-balancing designs of Qwen3~\citep{qwen3technicalreport} and Kimi K3~\citep{kimi2026k3}, our MoE routers combine an auxiliary load-balancing loss with auxiliary-loss-free expert-bias updates. We do not use a router z-loss. Expert dispatch has no capacity limit and uses no capacity factor, so no tokens are dropped.

\textbf{MoD layer hierarchy.}
For the MoD baseline, we follow the original one-sparse--one-dense recommendation with sparsity ratio \(1{:}8\)~\cite{MoD}. To match the 28-layer active-equivalent dense backbone, we use
\begin{equation}
[Dense] + ([Sparse(8)] + [Dense])_{\times 24},
\label{eq:appendix_mod_structure}
\end{equation}
which gives \(1+24+24/8=28\) active-equivalent layers and 49 total layers.

\textbf{X-MoD layer hierarchy.}
For X-MoD, we choose the layer hierarchy so that all compared variants share the same active-equivalent backbone size \(N_{\mathrm{act}}\). The X-MoD A4K8 configuration uses one shorter sparse--dense stage followed by four A4K8 stages:
\begin{equation}
[Dense]_{\times 4}
+
\left([Sparse(8)]_{\times (3\cdot 8)} + [Dense]\right)
+
\left([Sparse(8)]_{\times (4\cdot 8)} + [Dense]\right)_{\times 4},
\label{eq:appendix_xmod_a4k8_structure}
\end{equation}
this yields \(4+5+(3\cdot8+4\cdot4\cdot8)/8=28\) active-equivalent layers and 161 total layers. The X-MoD A3K16 configuration uses
\begin{equation}
[Dense]_{\times 4}
+
\left([Sparse(16)]_{\times (3\cdot 16)} + [Dense]\right)_{\times 6},
\label{eq:appendix_xmod_a3k16_structure}
\end{equation}
which gives \(4+6+(6\cdot3\cdot16)/16=28\) active-equivalent layers and 298 total layers. These constructions instantiate the same total-vs-active decoupling defined in the main text, while keeping the active-equivalent budget matched across Dense, MoD, MoE, and X-MoD variants.

\begin{table}[ht]
\centering
\caption{\textbf{Backbone model configurations.} All models use GQA attention and standard feed-forward layers.}
\label{tab:backbone_configs}
\begin{tabular}{lcccccc}
\toprule
\(N_{\mathrm{act}}\) & Dim & Head dim & Layers & Heads & KV heads & FFN dim \\
\midrule
165M  & 512  & 64  & 28 & 16 & 8 & 1536 \\
298M  & 768  & 64  & 28 & 16 & 8 & 2304 \\
556M  & 1024 & 128 & 28 & 16 & 8 & 3072 \\
936M  & 1536 & 128 & 28 & 16 & 8 & 3840 \\
1.65B & 2048 & 128 & 28 & 16 & 8 & 6144 \\
\bottomrule
\end{tabular}
\end{table}

\textbf{FLOPs accounting.}
All comparisons are aligned by logged training FLOPs. The FLOPs counter uses the actual model configuration, including sequence length, width, attention type, FFN or MoE configuration, and sparse-layer routing fraction. For X-MoD, sparse layers are counted according to their routed token fraction \(1/K\). For MoE baselines, expert computation is counted according to the number of activated experts per token. The same FLOPs accounting implementation is used for Dense, MoD, X-MoD, and MoE runs.

\textbf{Optimizer and learning-rate selection.}
We compare AdamW and Muon on dense baselines. For each optimizer, we sweep eight geometrically spaced peak learning rates:
\begin{equation}
\left\{ 2^0, 2^1, 2^2, 2^3, 2^4, 2^5, 2^6, 2^7\right\} * 10^{-4}.
\end{equation}

Muon with peak learning rate \(3.2\times10^{-3}\) gives the best dense validation loss and is used for the main runs. The Muon hyperparameters are listed in Table~\ref{tab:muon_config}.

\begin{table}[ht]
\centering
\caption{Muon optimizer configuration selected in preliminary dense-baseline sweeps and used in the main runs.}
\label{tab:muon_config}
\begin{tabular}{lr}
\toprule
Hyperparameter & Value \\
\midrule
Peak learning rate & \(3.2\times10^{-3}\) \\
Momentum coefficient \(\beta\) & \(0.95\) \\
Newton--Schulz steps & \(5\) \\
Nesterov momentum & True \\
RMS match & \(0.2\) \\
AdamW betas & \((0.9, 0.95)\) \\
AdamW epsilon & \(10^{-8}\) \\
Weight decay & \(0.1\) \\
\bottomrule
\end{tabular}
\end{table}

\textbf{Validation budget.}
Validation language-modeling loss is computed on approximately \(0.2\)B held-out tokens. All models are evaluated on the same held-out shards, making loss comparisons paired across architectures. For main comparisons and key ablations, larger held-out evaluations can be used to verify small loss gaps when necessary.

\textbf{Downstream evaluation.}
All downstream evaluations are zero-shot and use \texttt{lm-evaluation-harness}~\citep{eval-harness}. We evaluate selected checkpoints on HellaSwag~\cite{benchmark-1-hellaswag}, ARC-Easy~\cite{benchmark-2-arc}, ARC-Challenge~\cite{benchmark-2-arc}, PIQA~\cite{benchmark-3-piqa}, LAMBADA~\cite{benchmark-4-lambda} and BoolQ~\cite{benchmark-5-boolq}. These tasks are used as transfer sanity checks for pretraining quality, and no architecture-specific downstream hyperparameters are tuned.

\textbf{Compute resources.}
Experiments were conducted on NVIDIA A100 80GB GPUs using PyTorch DDP, except for the 1.65B MoE 8:128 and X-MoD A3K16 configurations, which use TP=8 as reported in Table~\ref{tab:main_model_configurations}. 
The largest experiments used up to 8 nodes with 8 GPUs per node. 
All model comparisons in the main text are aligned by logged training FLOPs rather than wall-clock time, since different routing configurations have different per-step costs.

\subsection{Main-comparison protocol}
\label{sec:methods_main_comparison}

The main comparisons use total training-compute budgets of \(C=10^{20}\), \(1.5\times10^{20}\) and \(2.5\times10^{20}\) FLOPs at \(N_{\mathrm{act}}=556\mathrm{M}\), \(936\mathrm{M}\) and \(1.65\mathrm{B}\), respectively.
All models use \(L=32768\); within each scale, Dense, MoD, MoE and X-MoD are compared at matched total training compute.
Detailed model and optimizer configurations are provided in Tables~\ref{tab:main_model_configurations} and \ref{tab:muon_config}, respectively.
We evaluate the corresponding checkpoints on HellaSwag, ARC-Easy, ARC-Challenge, PIQA, LAMBADA and BoolQ using the same zero-shot evaluation protocol.
Validation losses, task scores and their unweighted mean are reported to four significant figures, with scores expressed as percentages.
In Table~\ref{tab:main_model_configurations}, Experts is formatted as active routed experts : total routed experts + shared experts; \(d_e\) is the expert FFN dimension.
FLOPs per token are reported at \(L=32768\), and Exec.\ denotes DDP or eight-way tensor parallelism (TP=8).

\begin{table}[!p]
\centering
\caption{
\textbf{Detailed model configurations used in the main comparison.}
\(N_{\mathrm{act}}\) denotes active-equivalent parameters, \(N\) denotes total parameters and \(C\) denotes total training FLOPs.
}
\label{tab:main_model_configurations}
\label{tab:model_details}
\footnotesize
\setlength{\tabcolsep}{1.4pt}
\renewcommand{\arraystretch}{1.06}
\begin{tabular*}{\textwidth}{@{\extracolsep{\fill}}lllrlrrrr@{}}

\toprule
\(N_{\mathrm{act}}\) & \(C\) & Model & Layers & Experts & \(d_e\) & \(N\) & FLOPs/tok. & Exec. \\
\midrule
\multirow{6}{*}{556M} & \multirow{6}{*}{\(10^{20}\)}
& Dense         & 28  & --       & --  & 0.54B & \(8.4\times10^9\) & DDP \\
& & MoD           & 49  & --       & --  & 0.87B & \(7.7\times10^9\) & DDP \\
& & MoE 8:64      & 28  & 6:64+2   & 384 & 2.46B & \(8.4\times10^9\) & DDP \\
& & X-MoD A4K8    & 161 & --       & --  & 2.64B & \(3.9\times10^9\) & DDP \\
& & MoE 8:128     & 28  & 6:128+2  & 384 & 4.58B & \(8.4\times10^9\) & DDP \\
& & X-MoD A3K16   & 298 & --       & --  & 4.79B & \(3.9\times10^9\) & DDP \\
\midrule
\multirow{6}{*}{936M} & \multirow{6}{*}{\(1.5\times10^{20}\)}
& Dense         & 28  & --       & --  & 936M  & \(9.0\times10^9\) & DDP \\
& & MoD           & 49  & --       & --  & 1.48B & \(8.3\times10^9\) & DDP \\
& & MoE 8:64      & 28  & 6:64+2   & 480 & 4.51B & \(9.0\times10^9\) & DDP \\
& & X-MoD A4K8    & 161 & --       & --  & 4.52B & \(4.6\times10^9\) & DDP \\
& & MoE 8:128     & 28  & 6:128+2  & 480 & 8.48B & \(9.0\times10^9\) & DDP \\
& & X-MoD A3K16   & 298 & --       & --  & 8.24B & \(4.5\times10^9\) & DDP \\
\midrule
\multirow{6}{*}{1.65B} & \multirow{6}{*}{\(2.5\times10^{20}\)}
& Dense         & 28  & --       & --  & 1.65B  & \(1.0\times10^{10}\) & DDP \\
& & MoD           & 49  & --       & --  & 2.67B  & \(9.6\times10^9\)  & DDP \\
& & MoE 8:64      & 28  & 6:64+2   & 768 & 9.28B  & \(1.0\times10^{10}\) & DDP \\
& & X-MoD A4K8    & 161 & --       & --  & 8.31B  & \(5.9\times10^9\)  & DDP \\
& & MoE 8:128     & 28  & 6:128+2  & 768 & 17.16B & \(1.0\times10^{10}\) & TP=8 \\
& & X-MoD A3K16   & 298 & --       & --  & 14.73B & \(5.8\times10^9\)  & TP=8 \\
\bottomrule
\end{tabular*}
\end{table}

\subsection{Pilot configuration-selection protocol}
\label{sec:methods_pilot_selection}

\begin{figure}[!p]
\centering
\includegraphics[width=0.75\textwidth]{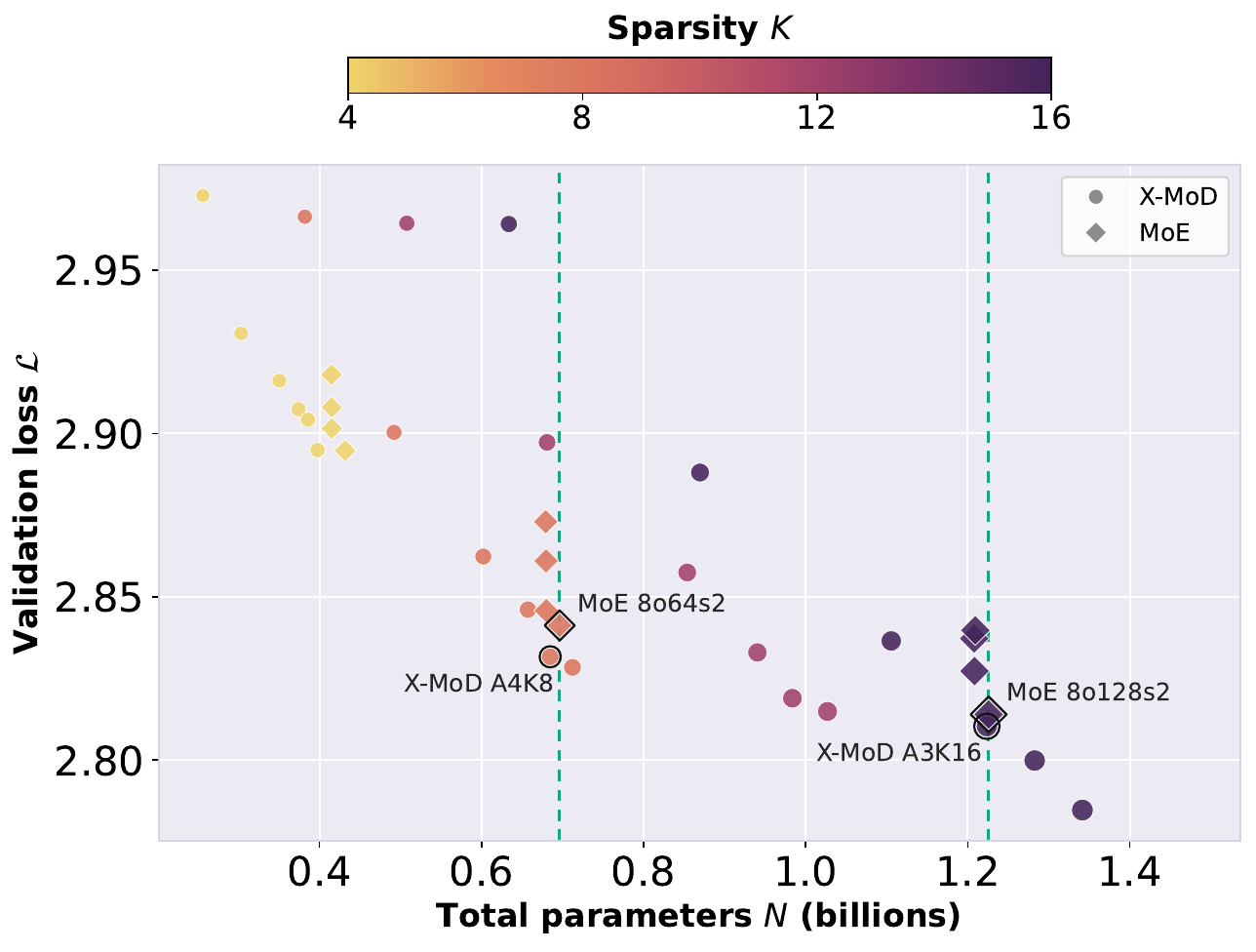}
\caption{
\textbf{Pilot comparison of X-MoD and MoE configurations.}
Validation loss versus total parameters at \(N_{\mathrm{act}}=165\mathrm{M}\), \(L=32\mathrm{k}\) and \(C=3\times10^{19}\) FLOPs.
Circles denote X-MoD and diamonds denote MoE; colour indicates \(K\), and marker area increases with total parameters.
Green dashed lines mark the selected MoE parameter budgets; black outlines identify the selected MoE--X-MoD pairs.
The plot labels 8o64s2 and 8o128s2 correspond to MoE 8:64 and 8:128.
}
\label{fig:xmod_moe_pilot_grid}
\end{figure}

The matched-budget pilot in Fig.~\ref{fig:xmod_moe_pilot_grid} compares 24 X-MoD configurations spanning \(A\in\{0.5,1,2,3,4,5\}\) and \(K\in\{4,8,12,16\}\) with 12 MoE configurations spanning \(K\in\{4,8,16\}\), expert granularity \(G\in\{2,4,8\}\), and either no shared experts or, for \(G=8\), two shared experts.
For MoE, \(K\) denotes the nominal ratio in the configuration name (e.g. \(64/8=8\)); the active-expert count includes shared experts.
The representative MoE and X-MoD settings are selected using this pilot sweep before the larger-scale comparisons.

\subsection{Scaling-law sweeps and out-of-range protocol}
\label{sec:methods_scaling_sweeps}

Figure~\ref{fig:ScalingLaw} summarizes the sweeps used to motivate and fit the law.
For the \(K\)- and \(L\)-sweeps, we vary sparsity across multiple context lengths at fixed \(N_{\mathrm{act}}=556\mathrm{M}\) and \(A=1\), comparing losses at matched compute within each \(L\).
The in-range context grid is \(L\in\{2\mathrm{k},4\mathrm{k},8\mathrm{k},16\mathrm{k},32\mathrm{k}\}\).
We additionally evaluate \(K=3\) at \(L=2\mathrm{k}\) to localize its shallow minimum more precisely; this extra \(K\) value is not used for the other context lengths.

Before fitting, we designate \(L=64\mathrm{k}\) as an out-of-range extension beyond the largest context used to construct the law.
At \(64\mathrm{k}\), we evaluate \(K\in\{1,16,18,20,22,24\}\); \(K=1\) provides the dense reference, and all of these measurements are excluded from fitting the multivariable law and are shown as squares with a dashed descriptive curve.
Thus, out-of-range denotes extrapolation beyond the predefined fitting domain, rather than a point removed post hoc from within that domain.

For the scale sweep, we vary \(N_{\mathrm{act}}\) at fixed \(L=32\mathrm{k}\) and \(A=1\), comparing each sparse configuration with its dense reference at matched training FLOPs.
The in-range active-equivalent parameter counts are \(\{165\mathrm{M},298\mathrm{M},556\mathrm{M},936\mathrm{M}\}\).
The \(N_{\mathrm{act}}=1.65\mathrm{B}\) setting is a prespecified out-of-range extension, evaluated at \(K\in\{14,16,18,20\}\) and excluded from fitting the multivariable law.
For the anchor-stride sweep, we vary \(A\in\{0.5,1,2,3,4,5\}\) and \(K\in\{4,8,12,16\}\) at fixed \(L=32\mathrm{k}\), \(N_{\mathrm{act}}=165\mathrm{M}\). %
The stars in Fig.~\ref{fig:ScalingLaw} mark the lowest measured loss on each evaluated grid; in particular, \(\hat A=5\) is the best tested anchor stride and does not assert a minimum beyond the evaluated range.

Each in-range experiment contributes one validation-loss observation at its target training-compute budget.
The smooth curves in Fig.~\ref{fig:ScalingLaw} are one-dimensional descriptive fits used to visualize each sweep; they are not predictions from Eq.~\eqref{eq:xmod_final_law}.

\subsection{Fitting and out-of-sample validation protocol}
\label{app:oos_protocol}
\label{sec:methods_scaling_validation}

We obtain \(\widehat{\mathcal L}_{\mathrm{dense}}(C,L,N_{\mathrm{act}})\) by interpolating dense validation curves at the same logged FLOPs for the corresponding sequence length and backbone family.

We use grouped held-out splits rather than random point-level splits.
Points within the same sweep are highly correlated, so randomly holding out individual points would mostly test interpolation within an already observed sweep.
In contrast, grouped splits evaluate whether the law transfers across unseen context lengths, anchor strides or model scales.
For leave-one-context-out, we hold out all sparse runs with one sequence length \(L\), refit the base law on the remaining context-length groups, and evaluate it on the held-out \(L\).
For leave-one-scale-out, we analogously hold out all sparse runs at one active-equivalent parameter count \(N_{\mathrm{act}}\), refit on the remaining scale groups, and evaluate on the excluded scale.

For leave-one-anchor-stride-out, we follow the same two-stage fitting procedure used for the all-data reduced law.
We first fit the base residual law on the \(A=1\) context-length and model-scale sweeps, and then calibrate the \(A\)-interaction using the controlled anchor-stride sweep with one non-baseline \(A\) group removed.
The \(A=1\) observations define the zero-interaction baseline because \(R_{A,K}=0\) at \(A=1\); they are therefore not treated as a held-out anchor-interaction group.
At each stage, the residual offset is unconstrained, the reward and correction coefficients are constrained to be non-negative, and these linear coefficients are solved conditionally for each candidate set of nonlinear exponents.
We optimize the nonlinear exponents by differential evolution to minimize mean squared residual error and solve the conditional linear problem by constrained least squares, using the same procedure for the all-data reduced-law fit and every grouped refit.

For the all-data reduced-law fit, we summarize descriptive goodness of fit using the unweighted coefficient of determination on matched-dense residuals,
\(R_{\Delta}^{2}=1-\sum_i(\Delta_i-\widehat{\Delta}_i)^2/\sum_i(\Delta_i-\overline{\Delta})^2\),
over the 109 in-range observations used for fitting.
The prespecified out-of-range observations are excluded from this statistic.

The out-of-range evaluation uses a single law fitted to all in-range observations and freezes its parameters before prediction at \(L=64\mathrm{k}\) and \(N_{\mathrm{act}}=1.65\mathrm{B}\).
No sparse configuration from either out-of-range setting is used to fit the law.
To report raw validation loss, we add the predicted residual to the matched dense reference (or to the shared dense level for the controlled \(A\)-sweep):
\begin{equation}
\widehat{\mathcal L}_i
=
\widehat{\mathcal L}_{\mathrm{dense},i}
+
\widehat\Delta_i.
\label{eq:raw_loss_reconstruction}
\end{equation}
Dense observations at a held-out or out-of-range setting are used only to define this matched-FLOPs reference and are not used to fit the sparse-depth residual law.

We assess raw-loss calibration using root mean squared error and mean absolute error:
\begin{equation}
\begin{aligned}
\mathrm{RMSE}_{\mathcal L}
&=
\sqrt{\frac{1}{n}\sum_{i=1}^{n}
\left(\widehat{\mathcal L}_i-\mathcal L_i\right)^2},\\
\mathrm{MAE}_{\mathcal L}
&=
\frac{1}{n}\sum_{i=1}^{n}
\left|\widehat{\mathcal L}_i-\mathcal L_i\right|.
\end{aligned}
\label{eq:raw_loss_metrics}
\end{equation}
For Fig.~\ref{fig:scaling_validation}a, the metrics are computed separately over the in-range configurations and over the nine out-of-range configurations pooled across \(L=64\mathrm{k}\) and \(N_{\mathrm{act}}=1.65\mathrm{B}\).
For each grouped held-out panel, they are pooled over all predictions produced when each complete group is withheld in turn.
In Fig.~\ref{fig:scaling_validation}, the diagonal denotes exact prediction and the shaded band marks an absolute loss error of 0.02; it is a visual guide, not a confidence interval.
Axes are scaled independently across panels, with identical horizontal and vertical limits within each panel.

\subsection{Ablation and controlled-scaling protocols}
\label{sec:methods_ablations}

\textbf{Mechanism controls.}
The Routed-FFN and Selected-Q/full-KV controls use the 936M reference backbone, \(L=32768\), the same held-out validation shards, and the common \(C=1.5\times10^{20}\)-FLOP checkpoint.
Between dense anchors, Routed-FFN follows \(\bigl([\mathrm{FullAttn}]+[\mathrm{Routed\mbox{-}FFN}]_{\times K}\bigr)_{\times A}\), where FullAttn denotes full-sequence causal attention and only the FFN refinements are routed.
The model retains 28 full-sequence attention modules in total.
The A4K8 and A3K16 variants contain 152 and 288 routed FFN refinements, respectively; both have \(\phi/\phi_D=1.00\), with total parameter counts of 3.26B and 5.68B.
Selected-Q/full-KV retains the corresponding X-MoD router, selected queries, total depth and total parameters, but uses all tokens as keys and values in each sparse layer.
This diagnostic is not strictly active-equivalent matched because every token activates the K/V projections.
At fixed training compute, Selected-Q/full-KV processes 46.8\% and 42.0\% of the training tokens processed by full X-MoD at A4K8 and A3K16, respectively, because of its higher per-token FLOPs.
The comparison tests the complete architecture under a fixed compute budget; it does not isolate the effect of full context at equal token exposure.
Table~\ref{tab:counterfactual_configurations} summarizes the control configurations; Routed-FFN depth counts FFN layers.

\textbf{Component removals.}
We use X-MoD A1K16 as the full-model reference and remove one component at a time.
The component-removal experiments in Table~\ref{tab:xmod_component_ablations} use \(N_{\mathrm{act}}=936\)M, \(L=32768\), and \(C=1.5\times10^{20}\) FLOPs.
\(\Delta\)Loss is measured relative to the full X-MoD model.
The same six downstream tasks are evaluated for the full model and each component removal, with task-level results included in the same table.

\textbf{Deep-model sweep.}
These models use the width configuration of the 165M backbone: \(d=512\), head dimension 64, 16 query heads, 8 KV heads and FFN dimension 1536.
Increasing the active-equivalent depth from 28 to 52 layers gives \(N_{\mathrm{act}}=256\)M.
We evaluate Deep Dense and Deep X-MoD A1K4, A1K8, A1K12 and A1K16 at \(C=4\times10^{19}\) FLOPs.
All Deep X-MoD variants use dense anchors, gated residual scaling and token-choice bias.
Because \(K\), total parameters, per-token FLOPs and training-token exposure vary jointly, we treat this as matched-active-capacity and matched-compute total-depth scaling rather than a single-factor depth ablation.
\begin{table}[!p]
\centering
\caption{
\textbf{Mechanism-control configurations at the 936M reference scale.}
\(D/D_D=(\phi/\phi_D)^{-1}\) denotes token exposure relative to Dense.
}
\label{tab:counterfactual_configurations}
\footnotesize
\setlength{\tabcolsep}{4pt}
\renewcommand{\arraystretch}{1.08}
\begin{tabular*}{\textwidth}{@{\extracolsep{\fill}}lll@{}}
\multicolumn{3}{@{}l}{\textbf{a, Routed computation and attention context}}\\[-1pt]
\toprule
Model & Routed computation & KV context \\
\midrule
Dense                       & None            & Full                         \\
MoD                         & Attention + FFN & Selected subset              \\
Routed-FFN A4K8             & FFN only        & Full every \(K\) refinements \\
Routed-FFN A3K16            & FFN only        & Full every \(K\) refinements \\
Selected-Q/full-KV A4K8     & Q/O + FFN       & Full sequence                \\
Selected-Q/full-KV A3K16    & Q/O + FFN       & Full sequence                \\
X-MoD A4K8                  & Attention + FFN & Selected subset              \\
X-MoD A3K16                 & Attention + FFN & Selected subset              \\
\bottomrule
\end{tabular*}

\vspace{10pt}
\begin{tabular*}{\textwidth}{@{\extracolsep{\fill}}lrrrr@{}}
\multicolumn{5}{@{}l}{\textbf{b, Depth, capacity, compute and relative token exposure}}\\[-1pt]
\toprule
Model & Depth & \(N\) & \(\phi/\phi_D\) & \(D/D_D\) \\
\midrule
Dense                       & 28  & 936M  & 1.00 & 1.00 \\
MoD                         & 49  & 1.48B & 0.92 & 1.09 \\
Routed-FFN A4K8             & 161 & 3.26B & 1.00 & 1.00 \\
Routed-FFN A3K16            & 298 & 5.68B & 1.00 & 1.00 \\
Selected-Q/full-KV A4K8     & 161 & 4.52B & 1.09 & 0.92 \\
Selected-Q/full-KV A3K16    & 298 & 8.24B & 1.19 & 0.84 \\
X-MoD A4K8                  & 161 & 4.52B & 0.51 & 1.96 \\
X-MoD A3K16                 & 298 & 8.24B & 0.50 & 2.00 \\
\bottomrule
\end{tabular*}
\end{table}

\subsection{Routing diagnostic protocol}
\label{sec:methods_routing_diagnostics}

\textbf{Top-\(k\)-to-threshold routing alignment.}
\label{app:routing_alignment}
Following MoD, MoD and X-MoD are trained with non-causal top-\(k\) routing, but evaluated with a deployable causal threshold rule. During evaluation, a token is routed through a sparse layer when its routing probability is at least \(0.5\). To align the training-time top-\(k\) decisions with this threshold rule, we add a binary cross-entropy auxiliary loss between the top-\(k\) selection target and the threshold prediction:
\begin{equation}
\mathcal L_{\mathrm{route}}
=
-\frac{1}{n}\sum_i
\left[
q_i\log p_i + (1-q_i)\log(1-p_i)
\right],
\end{equation}
where \(q_i\in\{0,1\}\) denotes whether token \(i\) is selected by the top-\(k\) router and \(p_i=\sigma(s_i)\) is the routing probability. This auxiliary objective encourages the top-\(k\) decision boundary to align with the threshold \(0.5\), reducing the gap between training-time routing and deployable causal evaluation.

We tune the auxiliary loss coefficient over
\begin{equation}
\{10^{-2},10^{-3},10^{-4},10^{-5}\}.
\end{equation}
A coefficient that is too large over-emphasizes the routing objective and degrades language-modeling loss, while a coefficient that is too small leaves a larger train--eval routing gap. We select \(10^{-4}\) based on training and validation language-modeling losses in preliminary experiments. Final-checkpoint mask agreement is reported in Table~\ref{tab:routing_diagnostics} under the diagnostic protocol below.

We collect routing diagnostics at the 936M reference scale with \(L=32768\), a per-device evaluation batch size of one, and the \(C=1.5\times10^{20}\)-FLOP checkpoints.
Coverage, consecutive-layer overlap and sparse-update counts are computed from threshold-driven forward passes over consecutive sparse-layer intervals bounded by dense layers.
A1K16 and A3K16 use 16- and 48-layer intervals, respectively; A4K8 statistics use only complete 32-layer intervals.
Coverage is the fraction of tokens selected at least once in an interval.
Consecutive-layer overlap is the Jaccard index of adjacent selected-token sets, averaged over pairs with a non-empty union within each sequence--interval observation, and then over observations with a defined mean.
Coverage is also averaged over sequence--interval observations; update-count distributions instead pool token--interval observations.

For mask agreement, we make separate top-\(k\)-driven forward passes and compute both the top-\(k\) and threshold masks from the same routing scores at each sparse layer.
Agreement is obtained by pooling matching binary decisions over all included token--layer positions, including both jointly selected and jointly skipped tokens; it does not compare two separately propagated routing trajectories.

\textbf{Reference selection schemes.}
Both analytical references use the nominal selection fraction \(p=1/K\) and sparse-interval length \(T\).
Repeated identical subsets have coverage \(p\), consecutive-layer Jaccard \(1\), and update-count probabilities \(1-p\) at zero and \(p\) at \(T\).
Independent random subsets have expected coverage \(1-(1-p)^T\), the large-sequence Jaccard approximation \(p/(2-p)\), and update counts distributed as \(\mathrm{Binomial}(T,p)\).
These are reference selection schemes, not trained models, and use nominal rather than measured threshold selection fractions.

\textbf{Distribution visualization.}
Each token contributes its number of selected sparse layers within an interval, and the distributions pool these token--interval observations over the full support from zero to \(T\).
Curves are shape-preserving cubic interpolants through the discrete probabilities, with auxiliary zero-valued endpoints at \(-0.5\) and \(T+0.5\) for display.
Integer heights preserve the observed probabilities; non-integer positions and filled areas have no probability interpretation.

\subsection{Systems measurement protocol}
\label{sec:methods_systems}

We measure end-to-end training throughput, peak training memory and validation-forward throughput at the 936M active-equivalent scale on one node with eight NVIDIA A100 80GB GPUs using PyTorch DistributedDataParallel (DDP).
All runs use \(L=32\mathrm{k}\), a per-GPU microbatch size of one and a global batch size of 80 sequences.
Training throughput is averaged over 2,000 optimization steps (approximately 5.24B training tokens), excluding evaluation and checkpoint-saving intervals.
Validation-forward throughput is averaged over four complete evaluations of the 0.2B-token validation set and includes routing, token movement, full-vocabulary projection and loss computation.
Accordingly, this measurement is a validation-forward benchmark rather than a KV-cache prefill or autoregressive decoding benchmark.

\section{Supplementary results}
\label{app:supplementary_results}

\subsection{Scaling-law fits and exposure comparison}
\label{app:xmod_fit_params}

Table~\ref{tab:scaling_law_parameters} reports the fitted parameters of the practical law in Eq.~\eqref{eq:xmod_final_law}.
The retained features \(R_N\), \(P_L\) and \(R_{A,K}\) are defined in Eqs.~\eqref{eq:xmod_phi_n}--\eqref{eq:xmod_phi_a}.
The two-stage constrained fit uses all 109 in-range observations and explains 98.5\% of their matched-dense residual variance (\(R_{\Delta}^{2}=0.9853\)).
This unweighted, in-sample statistic excludes the prespecified \(L=64\mathrm{k}\) and \(N_{\mathrm{act}}=1.65\mathrm{B}\) out-of-range observations, as does the fit.
The coefficients \(u\) and \(w\) absorb the parameter-count and token-count units of \(N_{\mathrm{act}}\) and \(L\), respectively.

\begin{table}[!ht]
\centering
\caption{
\textbf{Fitted parameters of the practical X-MoD scaling law.}
For ease of reference, Eq.~\eqref{eq:xmod_final_law} is reproduced above the estimates.
}
\label{tab:scaling_law_parameters}
\label{tab:xmod_fit_params}
\small
\setlength{\tabcolsep}{16pt}
\renewcommand{\arraystretch}{1.08}
\begin{tabular}{@{}llr@{}}
\multicolumn{3}{@{}c@{}}{%
\(\displaystyle
\begin{aligned}
\Delta(C,L,N_{\mathrm{act}},K,A)
&\approx
e
    -u\,N_{\mathrm{act}}^{-\eta_N}
\left[
1-
\left(\frac{N(K,A)}{N_{\mathrm{act}}}\right)^{-\rho_N}
\right]\\
&\quad
    +w\,L^{-\eta_L}(K^{\rho_L}-1)
-z\,(K^{\rho_K}-1)
\left[
\left(\frac{2A}{A+1}\right)^{\rho_A}-1
\right].
\end{aligned}
\)}\\
\addlinespace[6pt]
\toprule
Law component & Parameter & Estimate \\
\midrule
Residual offset
& \(e\) & 0.01848 \\
\addlinespace[2pt]
\multirow{3}{*}{Sparse-capacity reward}
& \(u\)       & 0.93519 \\
& \(\eta_N\)  & 0.07784 \\
& \(\rho_N\)  & 1.19538 \\
\addlinespace[2pt]
\multirow{3}{*}{Sparse-context correction}
& \(w\)       & 105.21108 \\
& \(\eta_L\)  & 0.92624 \\
& \(\rho_L\)  & 0.48494 \\
\addlinespace[2pt]
\multirow{3}{*}{Anchor-stride interaction}
& \(z\)       & 0.07455 \\
& \(\rho_K\)  & 0.79900 \\
& \(\rho_A\)  & 0.28293 \\
\bottomrule
\end{tabular}
\end{table}

\textbf{Comparison with the reduced law.}
We first use the \(A=1\) sweeps to study the capacity--context trade-off without anchor-stride interactions, and then use the controlled \(A\)-sweep to characterize these interactions.
Under this common protocol, we compare the full candidate law with its nested reduced form (\(v=0\)).
The reduced law achieves a lower combined in-range fitting RMSE (0.008758 versus 0.01349), supporting omission of \(P_D\) from the practical law.
Substituting the retained terms into Eq.~\eqref{eq:xmod_general_residual} yields the practical law in Eq.~\eqref{eq:xmod_final_law}.
This is an empirical parameterization rather than an unconstrained curve fit: each retained feature is tied to a sparse-depth mechanism and anchored to the corresponding dense reference point.

\subsection{Pilot configuration comparison}
\label{app:pilot_results}

Within the \(K=8\) and \(K=16\) groups, MoE 8:64 and 8:128 achieve the lowest loss among the tested MoE variants.
X-MoD A4K8 and A3K16 achieve the lowest loss among the tested X-MoD configurations at the corresponding \(K\) without exceeding the respective MoE total-parameter budget.

\subsection{Total-depth scaling}
\label{app:depth_results}

Table~\ref{tab:xmod_depth_scaling} reports the results.
All evaluated Deep X-MoD configurations improve on the Deep Dense reference, with validation loss decreasing monotonically across the tested configurations.

\begin{table}[htbp]
\centering
\caption{
\textbf{Total-depth scaling of X-MoD.}
All variants use the width of the 165M backbone, with matched active-equivalent capacity and training compute \(C=4\times10^{19}\).
}
\label{tab:xmod_depth_scaling}
\small
\setlength{\tabcolsep}{4pt}
\begin{tabular}{lrrrrr}
\toprule
Variant & Total \(N\) & \(N/N_{\mathrm{act}}\) & Total layers & \(\phi/\phi_D\) & Val. loss \(\downarrow\) \\
\midrule
Deep Dense & 256M & 1.00 & 52 & 1.00 & 3.084 \\
Deep X-MoD A1K4 & 552M & 2.16 & 124 & 0.67 & 3.047 \\
Deep X-MoD A1K8 & 939M & 3.67 & 220 & 0.62 & 3.034 \\
Deep X-MoD A1K12 & 1.29B & 5.04 & 316 & 0.60 & 3.023 \\
Deep X-MoD A1K16 & \textbf{1.67B} & \textbf{6.52} & \textbf{412} & \textbf{0.59} & \textbf{3.011} \\
\bottomrule
\end{tabular}
\end{table}

\subsection{Routing diagnostics}
\label{app:routing_results}

\begin{figure}[!htbp]
\centering
\includegraphics[width=\textwidth]{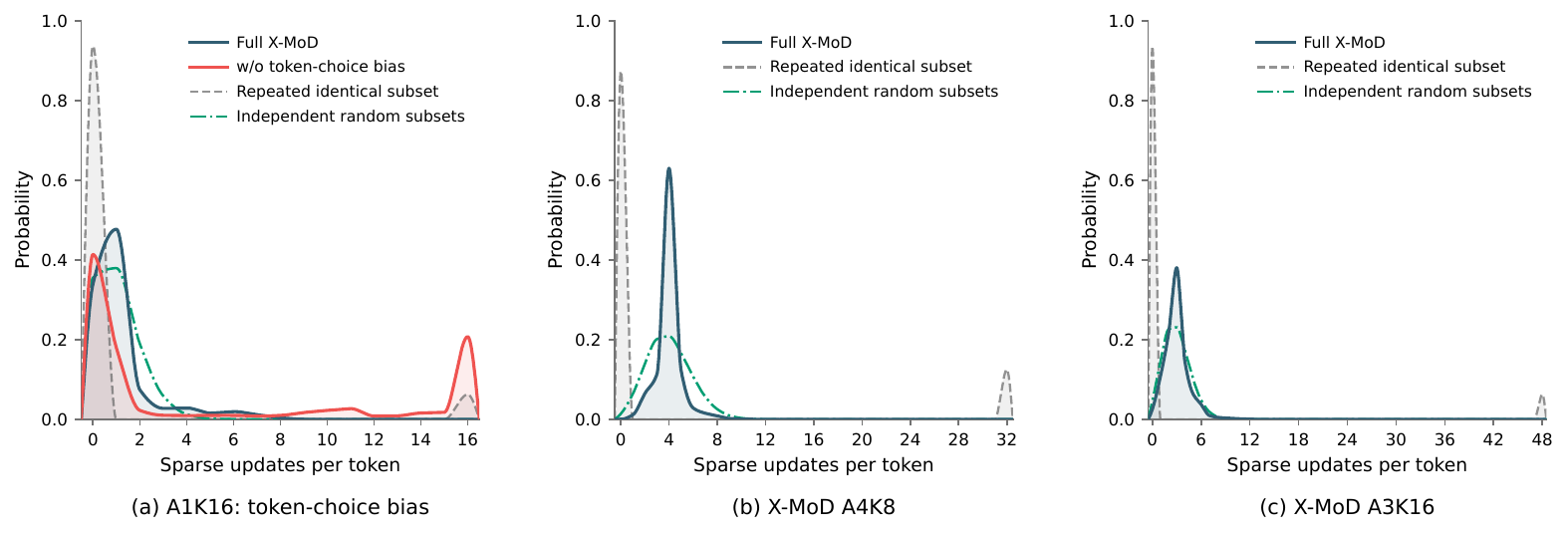}
\caption{
\textbf{Token-wise sparse-update distributions under threshold routing.}
(a) A1K16 with and without token-choice bias (16-layer intervals).
(b) A4K8 (complete 32-layer intervals). (c) A3K16 (48-layer intervals).
Grey dashed and green dot-dashed curves show repeated-subset and independent-subset references.
Curves interpolate discrete probabilities; integer heights, not filled areas, represent probability.
Models and evaluation conditions follow Table~\ref{tab:routing_diagnostics}.
}
\label{fig:routing_distributions}
\end{figure}

Table~\ref{tab:routing_diagnostics} reports coverage, overlap and mask agreement; Fig.~\ref{fig:routing_distributions} shows the sparse-update distributions.

\begin{table}[!p]
\centering
\caption{
\textbf{Routing coverage, consecutive-layer overlap and mask agreement.}
All entries are percentages at the 936M reference scale; a dash denotes an unmeasured value.
}
\label{tab:routing_diagnostics}
\footnotesize
\setlength{\tabcolsep}{3pt}
\renewcommand{\arraystretch}{1.2}
\begin{tabular*}{\textwidth}{@{\extracolsep{\fill}}lrrrrrrr@{}}
\toprule
 & \multicolumn{2}{c}{\shortstack{Repeated identical\\subset}} & \multicolumn{2}{c}{\shortstack{Independent random\\subsets}} & \multicolumn{2}{c}{\shortstack{Threshold\\routing}} & \multicolumn{1}{c}{\shortstack{Threshold /\\top-\(k\)}} \\
\cmidrule(lr){2-3}\cmidrule(lr){4-5}\cmidrule(lr){6-7}\cmidrule(lr){8-8}
Model & Coverage & Jaccard & Coverage & Jaccard & Coverage & Jaccard & Agreement \\
\midrule
X-MoD A1K16 & 6.250 & 100.0 & 64.39 & 3.226 & 52.24 & 14.51 & 98.63 \\
\shortstack[l]{A1K16 w/o\\token-choice bias} & 6.250 & 100.0 & 64.39 & 3.226 & 58.59 & 80.66 & -- \\
X-MoD A4K8 & 12.50 & 100.0 & 98.61 & 6.667 & 99.99 & 1.947 & 96.62 \\
X-MoD A3K16 & 6.250 & 100.0 & 95.49 & 3.226 & 97.50 & 2.940 & 98.25 \\
\bottomrule
\end{tabular*}
\end{table}

\subsection{Systems efficiency}
\label{app:systems_results}

Table~\ref{tab:systems_benchmark} reports the systems measurements obtained using the protocol in Appendix~\ref{sec:methods_systems}.

\begin{table}[!htbp]
\centering
\caption{
\textbf{Measured systems efficiency at the 936M active-equivalent scale.}
Speedups are relative to the adjacent, approximately total-parameter-matched MoE baseline.
}
\label{tab:systems_benchmark}

\footnotesize
\setlength{\tabcolsep}{5pt}
\renewcommand{\arraystretch}{1.06}
\begin{tabular*}{\textwidth}{@{\extracolsep{\fill}}lrrrr@{}}
\multicolumn{5}{@{}l}{\textbf{a, Model scale, counted compute and peak training memory}}\\[-1pt]
\toprule
Model & Layers & \(N\) & FLOPs/token \(\downarrow\) & Peak memory (GB/GPU) \(\downarrow\) \\
\midrule
Dense         & 28  & 936M  & 9.0G & 32.08 \\
MoD           & 49  & 1.48B & 8.3G & 46.62 \\
\addlinespace[2pt]
MoE 8:64      & 28  & 4.51B & 9.0G & 55.46 \\
X-MoD A4K8    & 161 & 4.52B & 4.6G & 55.72 \\
\addlinespace[2pt]
MoE 8:128     & 28  & 8.48B & 9.0G & 71.96 \\
X-MoD A3K16   & 298 & 8.24B & 4.5G & 73.33 \\
\bottomrule
\end{tabular*}

\vspace{10pt}
\footnotesize
\setlength{\tabcolsep}{4pt}
\begin{tabular*}{\textwidth}{@{\extracolsep{\fill}}lrrrr@{}}
\multicolumn{5}{@{}l}{\textbf{b, Measured throughput and pairwise speedup}}\\[-1pt]
\toprule
Model & Train throughput & Val.-forward throughput & Train speedup & Val.-forward speedup \\
& (M tok/s) \(\uparrow\) & (M tok/s) \(\uparrow\) & vs. MoE \(\uparrow\) & vs. MoE \(\uparrow\) \\
\midrule
Dense         & 0.041 & 0.226 &       &       \\
MoD           & 0.046 & 0.234 &       &       \\
\addlinespace[2pt]
MoE 8:64      & 0.027 & 0.176 & 1.00\(\times\) & 1.00\(\times\) \\
X-MoD A4K8    & 0.052 & 0.305 & \textbf{1.93\(\times\)} & \textbf{1.73\(\times\)} \\
\addlinespace[2pt]
MoE 8:128     & 0.023 & 0.153 & 1.00\(\times\) & 1.00\(\times\) \\
X-MoD A3K16   & 0.039 & 0.217 & \textbf{1.70\(\times\)} & \textbf{1.42\(\times\)} \\
\bottomrule
\end{tabular*}
\end{table}

%% file: main.bbl
\begin{thebibliography}{49}
\providecommand{\natexlab}[1]{#1}
\providecommand{\url}[1]{\texttt{#1}}
\expandafter\ifx\csname urlstyle\endcsname\relax
  \providecommand{\doi}[1]{doi: #1}\else
  \providecommand{\doi}{doi: \begingroup \urlstyle{rm}\Url}\fi

\bibitem[Abnar et~al.(2025)Abnar, Shah, Busbridge, El-Nouby, Susskind, and
  Thilak]{abnar2025}
Samira Abnar, Harshay Shah, Dan Busbridge, Alaaeldin El-Nouby, Joshua~M.
  Susskind, and Vimal Thilak.
\newblock Parameters vs {FLOP}s: Scaling laws for optimal sparsity for
  mixture-of-experts language models.
\newblock In \emph{Proceedings of the 42nd International Conference on Machine
  Learning}, pp.\  204--230, 2025.
\newblock URL \url{https://proceedings.mlr.press/v267/abnar25a.html}.

\bibitem[Ainslie et~al.(2023)Ainslie, Lee-Thorp, de~Jong, Zemlyanskiy, Lebron,
  and Sanghai]{GQA}
Joshua Ainslie, James Lee-Thorp, Michiel de~Jong, Yury Zemlyanskiy, Federico
  Lebron, and Sumit Sanghai.
\newblock {GQA}: Training generalized multi-query transformer models from
  multi-head checkpoints.
\newblock In \emph{Proceedings of the 2023 Conference on Empirical Methods in
  Natural Language Processing}, pp.\  4895--4901, 2023.
\newblock URL \url{https://aclanthology.org/2023.emnlp-main.298/}.

\bibitem[Bae et~al.(2025)Bae, Kim, Bayat, Kim, Ha, Schuster, Fisch,
  Harutyunyan, Ji, Courville, and Yun]{mor-1-mor}
Sangmin Bae, Yujin Kim, Reza Bayat, Sungnyun Kim, Jiyoun Ha, Tal Schuster, Adam
  Fisch, Hrayr Harutyunyan, Ziwei Ji, Aaron Courville, and Se-Young Yun.
\newblock Mixture-of-recursions: Learning dynamic recursive depths for adaptive
  token-level computation.
\newblock In \emph{Advances in Neural Information Processing Systems},
  volume~38, pp.\  96572--96617, 2025.
\newblock URL
  \url{https://proceedings.neurips.cc/paper_files/paper/2025/file/8b08bbf8b420faa6eeb4020720582ec7-Paper-Conference.pdf}.

\bibitem[Bisk et~al.(2020)Bisk, Zellers, Bras, Gao, and Choi]{benchmark-3-piqa}
Yonatan Bisk, Rowan Zellers, Ronan~Le Bras, Jianfeng Gao, and Yejin Choi.
\newblock Piqa: Reasoning about physical commonsense in natural language.
\newblock In \emph{Proceedings of the AAAI conference on artificial
  intelligence}, volume~34, pp.\  7432--7439, 2020.
\newblock URL \url{https://ojs.aaai.org/index.php/AAAI/article/view/6239/6095}.

\bibitem[Chen et~al.(2025)Chen, Shang, Zhang, Xie, Sheng, Liu, Wang, Sun, Wu,
  and Wang]{mor-2-inner-thinking-transformer}
Yilong Chen, Junyuan Shang, Zhenyu Zhang, Yanxi Xie, Jiawei Sheng, Tingwen Liu,
  Shuohuan Wang, Yu~Sun, Hua Wu, and Haifeng Wang.
\newblock Inner thinking transformer: Leveraging dynamic depth scaling to
  foster adaptive internal thinking.
\newblock In \emph{Proceedings of the 63rd Annual Meeting of the Association
  for Computational Linguistics (Volume 1: Long Papers)}, pp.\  28241--28259,
  2025.
\newblock URL \url{https://aclanthology.org/2025.acl-long.1369/}.

\bibitem[Clark et~al.(2022)Clark, de~las Casas, Guy, Mensch, Paganini,
  Hoffmann, Damoc, Hechtman, Cai, Borgeaud, van~den Driessche, Rutherford,
  Hennigan, Johnson, Millican, Cassirer, Jones, Buchatskaya, Budden, Sifre,
  Osindero, Vinyals, Rae, Elsen, Kavukcuoglu, and Simonyan]{clark2022}
Aidan Clark, Diego de~las Casas, Aurelia Guy, Arthur Mensch, Michela Paganini,
  Jordan Hoffmann, Bogdan Damoc, Blake Hechtman, Trevor Cai, Sebastian
  Borgeaud, George van~den Driessche, Eliza Rutherford, Tom Hennigan, Matthew
  Johnson, Katie Millican, Albin Cassirer, Chris Jones, Elena Buchatskaya,
  David Budden, Laurent Sifre, Simon Osindero, Oriol Vinyals, Jack Rae, Erich
  Elsen, Koray Kavukcuoglu, and Karen Simonyan.
\newblock Unified scaling laws for routed language models.
\newblock In \emph{Proceedings of the 39th International conference on machine
  learning}, pp.\  4057--4086, 2022.
\newblock URL \url{https://proceedings.mlr.press/v162/clark22a.html}.

\bibitem[Clark et~al.(2019)Clark, Lee, Chang, Kwiatkowski, Collins, and
  Toutanova]{benchmark-5-boolq}
Christopher Clark, Kenton Lee, Ming-Wei Chang, Tom Kwiatkowski, Michael
  Collins, and Kristina Toutanova.
\newblock {B}ool{Q}: Exploring the surprising difficulty of natural yes/no
  questions.
\newblock In \emph{Proceedings of the 2019 Conference of the North {A}merican
  Chapter of the Association for Computational Linguistics: Human Language
  Technologies, Volume 1 (Long and Short Papers)}, pp.\  2924--2936, 2019.
\newblock URL \url{https://aclanthology.org/N19-1300/}.

\bibitem[Clark et~al.(2018)Clark, Cowhey, Etzioni, Khot, Sabharwal, Schoenick,
  and Tafjord]{benchmark-2-arc}
Peter Clark, Isaac Cowhey, Oren Etzioni, Tushar Khot, Ashish Sabharwal, Carissa
  Schoenick, and Oyvind Tafjord.
\newblock Think you have solved question answering? try arc, the ai2 reasoning
  challenge, 2018.
\newblock URL \url{https://arxiv.org/abs/1803.05457}.

\bibitem[Dai et~al.(2024)Dai, Deng, Zhao, Xu, Gao, Chen, Li, Zeng, Yu, Wu, Xie,
  Li, Huang, Luo, Ruan, Sui, and Liang]{DeepSeekMoE2024}
Damai Dai, Chengqi Deng, Chenggang Zhao, R.x. Xu, Huazuo Gao, Deli Chen, Jiashi
  Li, Wangding Zeng, Xingkai Yu, Y.~Wu, Zhenda Xie, Y.k. Li, Panpan Huang, Fuli
  Luo, Chong Ruan, Zhifang Sui, and Wenfeng Liang.
\newblock {D}eep{S}eek{M}o{E}: Towards ultimate expert specialization in
  mixture-of-experts language models.
\newblock In \emph{Proceedings of the 62nd Annual Meeting of the Association
  for Computational Linguistics (Volume 1: Long Papers)}, pp.\  1280--1297,
  2024.
\newblock URL \url{https://aclanthology.org/2024.acl-long.70/}.

\bibitem[Du et~al.(2022)Du, Huang, Dai, Tong, Lepikhin, Xu, Krikun, Zhou, Yu,
  Firat, Zoph, Fedus, Bosma, Zhou, Wang, Wang, Webster, Pellat, Robinson,
  Meier-Hellstern, Duke, Dixon, Zhang, Le, Wu, Chen, and Cui]{GLaM2022}
Nan Du, Yanping Huang, Andrew~M Dai, Simon Tong, Dmitry Lepikhin, Yuanzhong Xu,
  Maxim Krikun, Yanqi Zhou, Adams~Wei Yu, Orhan Firat, Barret Zoph, Liam Fedus,
  Maarten~P Bosma, Zongwei Zhou, Tao Wang, Emma Wang, Kellie Webster, Marie
  Pellat, Kevin Robinson, Kathleen Meier-Hellstern, Toju Duke, Lucas Dixon, Kun
  Zhang, Quoc Le, Yonghui Wu, Zhifeng Chen, and Claire Cui.
\newblock Glam: Efficient scaling of language models with mixture-of-experts.
\newblock In \emph{Proceedings of the 39th International conference on machine
  learning}, pp.\  5547--5569. PMLR, 2022.
\newblock URL \url{https://proceedings.mlr.press/v162/du22c.html}.

\bibitem[Elhoushi et~al.(2024)Elhoushi, Shrivastava, Liskovich, Hosmer, Wasti,
  Lai, Mahmoud, Acun, Agarwal, Roman, Aly, Chen, and Wu]{skiplayer-1-layerskip}
Mostafa Elhoushi, Akshat Shrivastava, Diana Liskovich, Basil Hosmer, Bram
  Wasti, Liangzhen Lai, Anas Mahmoud, Bilge Acun, Saurabh Agarwal, Ahmed Roman,
  Ahmed Aly, Beidi Chen, and Carole-Jean Wu.
\newblock {L}ayer{S}kip: Enabling early exit inference and self-speculative
  decoding.
\newblock In \emph{Proceedings of the 62nd Annual Meeting of the Association
  for Computational Linguistics (Volume 1: Long Papers)}, pp.\  12622--12642,
  2024.
\newblock URL \url{https://aclanthology.org/2024.acl-long.681/}.

\bibitem[Fan et~al.(2025)Fan, Jiang, Li, Meng, Han, Shang, Sun, and
  Wang]{skiplayer-3-not-all-layers}
Siqi Fan, Xin Jiang, Xiang Li, Xuying Meng, Peng Han, Shuo Shang, Aixin Sun,
  and Yequan Wang.
\newblock Not all layers of llms are necessary during inference.
\newblock In \emph{Proceedings of the Thirty-Fourth International Joint
  Conference on Artificial Intelligence}, 2025.
\newblock URL \url{https://doi.org/10.24963/ijcai.2025/566}.

\bibitem[Fedus et~al.(2022)Fedus, Zoph, and Shazeer]{SwitchTransformer2022}
William Fedus, Barret Zoph, and Noam Shazeer.
\newblock Switch transformers: Scaling to trillion parameter models with simple
  and efficient sparsity.
\newblock \emph{Journal of Machine Learning Research}, 23\penalty0
  (120):\penalty0 1--39, 2022.
\newblock URL \url{https://www.jmlr.org/papers/v23/21-0998.html}.

\bibitem[Frey et~al.(2026)Frey, Shomali, Bashir, Berghaus, Koehler, and
  Ali]{mor-5-adaptive-loops}
Markus Frey, Behzad Shomali, Ali~Hamza Bashir, David Berghaus, Joachim Koehler,
  and Mehdi Ali.
\newblock Adaptive loops and memory in transformers: Think harder or know more?
\newblock In \emph{ICLR 2026 Workshop on Latent \& Implicit Thinking - Going
  Beyond CoT Reasoning}, 2026.
\newblock URL \url{https://openreview.net/forum?id=F87X9c107e#discussion}.

\bibitem[Gao et~al.(2024)Gao, Tow, Abbasi, Biderman, Black, DiPofi, Foster,
  Golding, Hsu, Le~Noac'h, Li, McDonell, Muennighoff, Ociepa, Phang, Reynolds,
  Schoelkopf, Skowron, Sutawika, Tang, Thite, Wang, Wang, and
  Zou]{eval-harness}
Leo Gao, Jonathan Tow, Baber Abbasi, Stella Biderman, Sid Black, Anthony
  DiPofi, Charles Foster, Laurence Golding, Jeffrey Hsu, Alain Le~Noac'h,
  Haonan Li, Kyle McDonell, Niklas Muennighoff, Chris Ociepa, Jason Phang,
  Laria Reynolds, Hailey Schoelkopf, Aviya Skowron, Lintang Sutawika, Eric
  Tang, Anish Thite, Ben Wang, Kevin Wang, and Andy Zou.
\newblock The language model evaluation harness, 07 2024.
\newblock URL \url{https://zenodo.org/records/12608602}.

\bibitem[Ge et~al.(2024)Ge, Zhang, Liu, Zhang, Han, and
  Gao]{sparse-attn-4-model-tells}
Suyu Ge, Yunan Zhang, Liyuan Liu, Minjia Zhang, Jiawei Han, and Jianfeng Gao.
\newblock Model tells you what to discard: Adaptive kv cache compression for
  llms.
\newblock In \emph{International Conference on Learning Representations}, 2024.
\newblock URL \url{https://openreview.net/forum?id=uNrFpDPMyo}.

\bibitem[Hoffmann et~al.(2022)Hoffmann, Borgeaud, Mensch, Buchatskaya, Cai,
  Rutherford, de~Las~Casas, Hendricks, Welbl, Clark, Hennigan, Noland,
  Millican, van~den Driessche, Damoc, Guy, Osindero, Simonyan, Elsen, Rae,
  Vinyals, and Sifre]{hoffmann2022}
Jordan Hoffmann, Sebastian Borgeaud, Arthur Mensch, Elena Buchatskaya, Trevor
  Cai, Eliza Rutherford, Diego de~Las~Casas, Lisa~Anne Hendricks, Johannes
  Welbl, Aidan Clark, Tom Hennigan, Eric Noland, Katie Millican, George van~den
  Driessche, Bogdan Damoc, Aurelia Guy, Simon Osindero, Karen Simonyan, Erich
  Elsen, Jack~W. Rae, Oriol Vinyals, and Laurent Sifre.
\newblock Training compute-optimal large language models.
\newblock In \emph{Proceedings of the 36th International Conference on Neural
  Information Processing Systems}, pp.\  30016--30030, 2022.
\newblock URL
  \url{https://proceedings.neurips.cc/paper_files/paper/2022/file/c1e2faff6f588870935f114ebe04a3e5-Paper-Conference.pdf}.

\bibitem[Jeddi et~al.(2026)Jeddi, Ciccone, and Taati]{mor-4-loopformer}
Ahmadreza Jeddi, Marco Ciccone, and Babak Taati.
\newblock Loopformer: Elastic-depth looped transformers for latent reasoning
  via shortcut modulation.
\newblock In \emph{International Conference on Learning Representations}, 2026.
\newblock URL \url{https://openreview.net/forum?id=RzYXb5YWBs}.

\bibitem[Jiang et~al.(2024)Jiang, Wang, Xie, Zhao, Zhang, Qian, and
  Lui]{skiplayer-4-d-llm}
Yikun Jiang, Huanyu Wang, Lei Xie, Hanbin Zhao, Chao Zhang, Hui Qian, and
  John~C.S. Lui.
\newblock D-llm: A token adaptive computing resource allocation strategy for
  large language models.
\newblock In \emph{Advances in Neural Information Processing Systems},
  volume~37, pp.\  1725--1749, 2024.
\newblock URL
  \url{https://proceedings.neurips.cc/paper_files/paper/2024/file/03469b1a66e351b18272be23baf3b809-Paper-Conference.pdf}.

\bibitem[Jin et~al.(2025)Jin, Zhu, Yuan, and YAN]{nullexpert-2-moe++}
Peng Jin, Bo~Zhu, Li~Yuan, and Shuicheng YAN.
\newblock Moe++: Accelerating mixture-of-experts methods with zero-computation
  experts.
\newblock In \emph{International Conference on Learning Representations}, 2025.
\newblock URL \url{https://openreview.net/forum?id=t7P5BUKcYv}.

\bibitem[Jolicoeur-Martineau(2025)]{mor-6-less-is-more}
Alexia Jolicoeur-Martineau.
\newblock Less is more: Recursive reasoning with tiny networks, 2025.
\newblock URL \url{https://arxiv.org/abs/2510.04871}.

\bibitem[Kaplan et~al.(2020)Kaplan, McCandlish, Henighan, Brown, Chess, Child,
  Gray, Radford, Wu, and Amodei]{kaplan2020}
Jared Kaplan, Sam McCandlish, Tom Henighan, Tom~B. Brown, Benjamin Chess, Rewon
  Child, Scott Gray, Alec Radford, Jeffrey Wu, and Dario Amodei.
\newblock Scaling laws for neural language models, 2020.
\newblock URL \url{https://arxiv.org/abs/2001.08361}.

\bibitem[Kim et~al.(2024)Kim, Kim, Kim, Castells, Choi, Shin, and
  Song]{skiplayer-2-shortened-llama}
Bo-Kyeong Kim, Geonmin Kim, Tae-Ho Kim, Thibault Castells, Shinkook Choi, Junho
  Shin, and Hyoung-Kyu Song.
\newblock Shortened {LLaMA}: Depth pruning for large language models with
  comparison of retraining methods.
\newblock \emph{arXiv preprint arXiv:2402.02834}, 2024.
\newblock URL \url{https://arxiv.org/abs/2402.02834}.

\bibitem[{Kimi Team}(2026)]{kimi2026k3}
{Kimi Team}.
\newblock {Kimi K3}: Open frontier intelligence, 2026.
\newblock URL \url{https://arxiv.org/abs/2607.24653}.

\bibitem[Krajewski et~al.(2024)Krajewski, Ludziejewski, Adamczewski, Pióro,
  Krutul, Antoniak, Ciebiera, Król, Odrzygóźdź, Sankowski, Cygan, and
  Jaszczur]{ludziejewski2024}
Jakub Krajewski, Jan Ludziejewski, Kamil Adamczewski, Maciej Pióro, Michał
  Krutul, Szymon Antoniak, Kamil Ciebiera, Krystian Król, Tomasz
  Odrzygóźdź, Piotr Sankowski, Marek Cygan, and Sebastian Jaszczur.
\newblock Scaling laws for fine-grained mixture of experts.
\newblock In \emph{Proceedings of the 41st International Conference on Machine
  Learning}, pp.\  33270--33288, 2024.
\newblock URL \url{https://proceedings.mlr.press/v235/ludziejewski24a.html}.

\bibitem[Lepikhin et~al.(2021)Lepikhin, Lee, Xu, Chen, Firat, Huang, Krikun,
  Shazeer, and Chen]{GShard2020}
Dmitry Lepikhin, HyoukJoong Lee, Yuanzhong Xu, Dehao Chen, Orhan Firat, Yanping
  Huang, Maxim Krikun, Noam Shazeer, and Zhifeng Chen.
\newblock Gshard: Scaling giant models with conditional computation and
  automatic sharding.
\newblock In \emph{International Conference on Learning Representations}, 2021.
\newblock URL \url{https://openreview.net/forum?id=qrwe7XHTmYb}.

\bibitem[Lin et~al.(2024)Lin, Shrivastava, Luo, Iyer, Lewis, Ghosh,
  Zettlemoyer, and Aghajanyan]{mllm-1-moma}
Xi~Victoria Lin, Akshat Shrivastava, Liang Luo, Srinivasan Iyer, Mike Lewis,
  Gargi Ghosh, Luke Zettlemoyer, and Armen Aghajanyan.
\newblock Moma: Efficient early-fusion pre-training with mixture of
  modality-aware experts, 2024.
\newblock URL \url{https://arxiv.org/abs/2407.21770}.

\bibitem[Lozhkov et~al.(2024)Lozhkov, Ben~Allal, von Werra, and
  Wolf]{finewebedu}
Anton Lozhkov, Loubna Ben~Allal, Leandro von Werra, and Thomas Wolf.
\newblock Fineweb-edu: the finest collection of educational content, 2024.
\newblock URL \url{https://huggingface.co/datasets/HuggingFaceFW/fineweb-edu}.

\bibitem[Ludziejewski et~al.(2025)Ludziejewski, Pióro, Krajewski, Stefaniak,
  Krutul, Małaśnicki, Cygan, Sankowski, Adamczewski, Miłoś, and
  Jaszczur]{jointmoescalinglaws2025}
Jan Ludziejewski, Maciej Pióro, Jakub Krajewski, Maciej Stefaniak, Michał
  Krutul, Jan Małaśnicki, Marek Cygan, Piotr Sankowski, Kamil Adamczewski,
  Piotr Miłoś, and Sebastian Jaszczur.
\newblock Joint moe scaling laws: Mixture of experts can be memory efficient.
\newblock In \emph{Proceedings of the 42nd International Conference on Machine
  Learning}, pp.\  41056--41073, 2025.
\newblock URL \url{https://proceedings.mlr.press/v267/ludziejewski25a.html}.

\bibitem[Luo et~al.(2025)Luo, Luo, Ji, Zhou, Sun, Shen, and
  Ji]{mllm-8-gamma-mod}
Yaxin Luo, Gen Luo, Jiayi Ji, Yiyi Zhou, Xiaoshuai Sun, Zhiqiang Shen, and
  Rongrong Ji.
\newblock $\gamma$-{MoD}: Exploring mixture-of-depth adaptation for multimodal
  large language models.
\newblock In \emph{International Conference on Learning Representations}, 2025.
\newblock URL \url{https://openreview.net/forum?id=q44uq3tc2D}.

\bibitem[Paperno et~al.(2016)Paperno, Kruszewski, Lazaridou, Pham, Bernardi,
  Pezzelle, Baroni, Boleda, and Fern{\'a}ndez]{benchmark-4-lambda}
Denis Paperno, Germ{\'a}n Kruszewski, Angeliki Lazaridou, Ngoc-Quan Pham,
  Raffaella Bernardi, Sandro Pezzelle, Marco Baroni, Gemma Boleda, and Raquel
  Fern{\'a}ndez.
\newblock The lambada dataset: Word prediction requiring a broad discourse
  context.
\newblock In \emph{Proceedings of the 54th annual meeting of the association
  for computational linguistics (volume 1: Long papers)}, pp.\  1525--1534,
  2016.

\bibitem[Pi{\k{e}}kos et~al.(2025)Pi{\k{e}}kos, Csord{\'a}s, and
  Schmidhuber]{sparse-attn-1-mixture-of-sparse-attention}
Piotr Pi{\k{e}}kos, R{\'o}bert Csord{\'a}s, and J{\"u}rgen Schmidhuber.
\newblock Mixture of sparse attention: Content-based learnable sparse attention
  via expert-choice routing.
\newblock In \emph{NeurIPS 2025 Workshop on Efficient Reasoning}, 2025.
\newblock URL \url{https://openreview.net/forum?id=6JEa0TKZWA}.

\bibitem[Raposo et~al.(2024)Raposo, Ritter, Richards, Lillicrap, Humphreys, and
  Santoro]{MoD}
David Raposo, Sam Ritter, Blake Richards, Timothy Lillicrap, Peter~Conway
  Humphreys, and Adam Santoro.
\newblock Mixture-of-depths: Dynamically allocating compute in
  transformer-based language models, 2024.
\newblock URL \url{https://arxiv.org/abs/2404.02258}.

\bibitem[Shazeer et~al.(2017)Shazeer, Mirhoseini, Maziarz, Davis, Le, Hinton,
  and Dean]{MoE2017}
Noam Shazeer, Azalia Mirhoseini, Krzysztof Maziarz, Andy Davis, Quoc Le,
  Geoffrey Hinton, and Jeff Dean.
\newblock Outrageously large neural networks: The sparsely-gated
  mixture-of-experts layer.
\newblock In \emph{International Conference on Learning Representations}, 2017.
\newblock URL \url{https://openreview.net/forum?id=B1ckMDqlg}.

\bibitem[Team(2024)]{team2024gemini}
Google~Gemini Team.
\newblock Gemini 1.5: Unlocking multimodal understanding across millions of
  tokens of context.
\newblock \emph{arXiv preprint arXiv:2403.05530}, 2024.
\newblock URL \url{https://arxiv.org/abs/2403.05530}.

\bibitem[Team(2025{\natexlab{a}})]{nullexpert-3-longcat}
Meituan~LongCat Team.
\newblock Longcat-flash technical report, 2025{\natexlab{a}}.
\newblock URL \url{https://arxiv.org/abs/2509.01322}.

\bibitem[Team(2025{\natexlab{b}})]{qwen3technicalreport}
Qwen Team.
\newblock Qwen3 technical report, 2025{\natexlab{b}}.
\newblock URL \url{https://arxiv.org/abs/2505.09388}.

\bibitem[Tian et~al.(2026)Tian, Chen, Liu, Liu, Zhang, and Zhou]{tian2025}
Changxin Tian, Kunlong Chen, Jia Liu, Ziqi Liu, Zhiqiang Zhang, and Jun Zhou.
\newblock Towards greater leverage: Scaling laws for efficient
  mixture-of-experts language models.
\newblock In \emph{International Conference on Learning Representations}, 2026.
\newblock URL \url{https://openreview.net/forum?id=7r2lkhDGUj}.

\bibitem[Wang et~al.(2025)Wang, Li, Sun, Chen, Liu, Wu, Lu, Song, and
  Yadkori]{mor-7-hierarchical-reasoning-model}
Guan Wang, Jin Li, Yuhao Sun, Xing Chen, Changling Liu, Yue Wu, Meng Lu, Sen
  Song, and Yasin~Abbasi Yadkori.
\newblock Hierarchical reasoning model, 2025.
\newblock URL \url{https://arxiv.org/abs/2506.21734}.

\bibitem[Wang et~al.(2024)Wang, Chen, Li, He, Zhang, and Wang]{wang2024}
Siqi Wang, Zhengyu Chen, Bei Li, Keqing He, Min Zhang, and Jingang Wang.
\newblock Scaling laws across model architectures: A comparative analysis of
  dense and {M}o{E} models in large language models.
\newblock In \emph{Proceedings of the 2024 Conference on Empirical Methods in
  Natural Language Processing}, pp.\  5583--5595, 2024.
\newblock URL \url{https://aclanthology.org/2024.emnlp-main.319/}.

\bibitem[Wu et~al.(2024)Wu, Chen, Lin, Wang, Gao, Xu, Xu, Hu, Chen, and
  Shou]{mllm-3-videollm-mod}
Shiwei Wu, Joya Chen, Kevin~Qinghong Lin, Qimeng Wang, Yan Gao, Qianli Xu, Tong
  Xu, Yao Hu, Enhong Chen, and Mike~Zheng Shou.
\newblock Videollm-mod: Efficient video-language streaming with
  mixture-of-depths vision computation.
\newblock In \emph{Advances in Neural Information Processing Systems},
  volume~37, pp.\  109922--109947, 2024.
\newblock URL
  \url{https://proceedings.neurips.cc/paper_files/paper/2024/file/c6a79e139ec4f371701ea8cc9e06018e-Paper-Conference.pdf}.

\bibitem[Xiao et~al.(2024)Xiao, Tian, Chen, Han, and
  Lewis]{sparse-attn-2-streamingllm}
Guangxuan Xiao, Yuandong Tian, Beidi Chen, Song Han, and Mike Lewis.
\newblock Efficient streaming language models with attention sinks.
\newblock In \emph{International Conference on Learning Representations}, 2024.
\newblock URL \url{https://openreview.net/forum?id=NG7sS51zVF}.

\bibitem[Xiong et~al.(2024)Xiong, Liu, Molybog, Zhang, Bhargava, Hou, Martin,
  Rungta, Sankararaman, Oguz, Khabsa, Fang, Mehdad, Narang, Malik, Fan,
  Bhosale, Edunov, Lewis, Wang, and Ma]{xiong2024effective}
Wenhan Xiong, Jingyu Liu, Igor Molybog, Hejia Zhang, Prajjwal Bhargava, Rui
  Hou, Louis Martin, Rashi Rungta, Karthik~Abinav Sankararaman, Barlas Oguz,
  Madian Khabsa, Han Fang, Yashar Mehdad, Sharan Narang, Kshitiz Malik, Angela
  Fan, Shruti Bhosale, Sergey Edunov, Mike Lewis, Sinong Wang, and Hao Ma.
\newblock Effective long-context scaling of foundation models.
\newblock In \emph{Proceedings of the 2024 Conference of the North American
  Chapter of the Association for Computational Linguistics: Human Language
  Technologies (Volume 1: Long Papers)}, pp.\  4643--4663, 2024.
\newblock URL \url{https://aclanthology.org/2024.naacl-long.260/}.

\bibitem[Yuan et~al.(2025)Yuan, Gao, Dai, Luo, Zhao, Zhang, Xie, Wei, Wang,
  Xiao, Wang, Ruan, Zhang, Liang, and Zeng]{sparse-attn-5-nsa}
Jingyang Yuan, Huazuo Gao, Damai Dai, Junyu Luo, Liang Zhao, Zhengyan Zhang,
  Zhenda Xie, Yuxing Wei, Lean Wang, Zhiping Xiao, Yuqing Wang, Chong Ruan,
  Ming Zhang, Wenfeng Liang, and Wangding Zeng.
\newblock Native sparse attention: Hardware-aligned and natively trainable
  sparse attention.
\newblock In Wanxiang Che, Joyce Nabende, Ekaterina Shutova, and Mohammad~Taher
  Pilehvar (eds.), \emph{Proceedings of the 63rd Annual Meeting of the
  Association for Computational Linguistics (Volume 1: Long Papers)}, pp.\
  23078--23097, 2025.
\newblock URL \url{https://aclanthology.org/2025.acl-long.1126/}.

\bibitem[Zellers et~al.(2019)Zellers, Holtzman, Bisk, Farhadi, and
  Choi]{benchmark-1-hellaswag}
Rowan Zellers, Ari Holtzman, Yonatan Bisk, Ali Farhadi, and Yejin Choi.
\newblock {H}ella{S}wag: Can a machine really finish your sentence?
\newblock In \emph{Proceedings of the 57th Annual Meeting of the Association
  for Computational Linguistics}, pp.\  4791--4800, 2019.
\newblock URL \url{https://aclanthology.org/P19-1472/}.

\bibitem[Zeng et~al.(2024)Zeng, Miao, Gao, Zhang, and
  Deng]{nullexpert-1-adamoe}
Zihao Zeng, Yibo Miao, Hongcheng Gao, Hao Zhang, and Zhijie Deng.
\newblock {A}da{M}o{E}: Token-adaptive routing with null experts for
  mixture-of-experts language models.
\newblock In \emph{Findings of the Association for Computational Linguistics:
  EMNLP 2024}, pp.\  6223--6235, 2024.
\newblock URL \url{https://aclanthology.org/2024.findings-emnlp.361/}.

\bibitem[Zhang et~al.(2025)Zhang, Meng, Zhang, Huang, Wu, and
  Wang]{mllm-9-p-mod}
Jun Zhang, Desen Meng, Zhengming Zhang, Zhenpeng Huang, Tao Wu, and Limin Wang.
\newblock p-mod: Building mixture-of-depths mllms via progressive ratio decay.
\newblock In \emph{Proceedings of the IEEE/CVF International Conference on
  Computer Vision}, pp.\  3705--3715, 2025.
\newblock URL
  \url{https://openaccess.thecvf.com/content/ICCV2025/papers/Zhang_p-MoD_Building_Mixture-of-Depths_MLLMs_via_Progressive_Ratio_Decay_ICCV_2025_paper.pdf}.

\bibitem[Zhang et~al.(2026)Zhang, Dong, Zhang, Heng, Chi, Dai, Du, Wang, Du,
  and Zhang]{mllm-2-mole-vla}
Rongyu Zhang, Menghang Dong, Yuan Zhang, Liang Heng, Xiaowei Chi, Gaole Dai,
  Li~Du, Dan Wang, Yuan Du, and Shanghang Zhang.
\newblock Mole-vla: Dynamic layer-skipping vision language action model via
  mixture-of-layers for efficient robot manipulation.
\newblock In \emph{Proceedings of the AAAI Conference on Artificial
  Intelligence}, volume~40, pp.\  18764--18772, 2026.
\newblock URL \url{https://ojs.aaai.org/index.php/AAAI/article/view/38945}.

\bibitem[Zhang et~al.(2023)Zhang, Sheng, Zhou, Chen, Zheng, Cai, Song, Tian,
  R\'{e}, Barrett, Wang, and Chen]{sparse-attn-3-h2o}
Zhenyu Zhang, Ying Sheng, Tianyi Zhou, Tianlong Chen, Lianmin Zheng, Ruisi Cai,
  Zhao Song, Yuandong Tian, Christopher R\'{e}, Clark Barrett,
  Zhangyang~"Atlas" Wang, and Beidi Chen.
\newblock H2o: Heavy-hitter oracle for efficient generative inference of large
  language models.
\newblock In \emph{Advances in Neural Information Processing Systems},
  volume~36, pp.\  34661--34710, 2023.
\newblock URL
  \url{https://proceedings.neurips.cc/paper_files/paper/2023/file/6ceefa7b15572587b78ecfcebb2827f8-Paper-Conference.pdf}.

\end{thebibliography}
